\documentclass{SBCbookchapter}
\usepackage[utf8]{inputenc} % latin1
\usepackage[T1]{fontenc}
\usepackage[brazilian]{babel}
\usepackage{graphicx}

\usepackage[cmex10]{amsmath}
\usepackage{amsthm, amsmath, amssymb}

\theoremstyle{plain}
\newtheorem{thm}{Teorema}[chapter] 
\usepackage{algorithmic}
\usepackage{listings}
\usepackage{array}
\usepackage{url,hyperref}
\usepackage{tikz}
\usetikzlibrary{positioning}
\usepackage{dblfloatfix}
\usepackage{varwidth}
\usepackage{fancyhdr}
\usepackage{xspace}
\usepackage{wrapfig}
\usepackage{placeins}
\usepackage{fancyhdr}
 
\usepackage[nodisplayskipstretch]{setspace}

\def\onedot{\ifx\@let@token.\else.\null\fi\xspace}

\def\etal{\emph{et al}\onedot}

\addtocounter{chapter}{2}
\title{Como funciona o \textit{Deep Learning}}
\author{Moacir A. Ponti$^1$ e Gabriel B. Paranhos da Costa\footnote{ICMC --- Universidade de São Paulo, São Carlos, SP, Brasil}}

\begin{document}

\maketitle

\begin{resumo}
    Métodos de Aprendizado Profundo (\textit{Deep Learning}) são atualmente o estado-da-arte em muitos problemas possíveis de se resolver via aprendizado de maquina, em particular problemas de classificação. No entanto, ainda há pouco entendimento de como esses métodos funcionam, porque funcionam e quais as limitações envolvidas ao utilizá-los. Nesse capítulo descreveremos em detalhes a transição desde redes rasas (\textit{shallow}) até as redes profundas (\textit{deep}), incluindo exemplos em código de como implementá-las, bem como os principais fatores a se levar em consideração ao treinar uma rede profunda. Adicionalmente, iremos introduzir aspectos teóricos que embasam o uso de modelos profundos, e discutir suas limitações.
\end{resumo}

\begin{abstract}
    Deep Learning methods are currently the state-of-the-art in many problems which can be tackled via machine learning, in particular classification problems. However there is still lack of understanding on how those methods work, why they work and what are the limitations involved in using them. In this chapter we will describe in detail the transition from shallow to deep networks, include examples of code on how to implement them, as well as the main issues one faces when training a deep network. Afterwards, we introduce some theoretical background behind the use of deep models, and discuss their limitations.
\end{abstract}

% \tableofcontents

\section{Introdução}
\label{sec:introduction}

Nos últimos anos, técnicas de Aprendizado Profundo tem revolucionado diversas áreas de aprendizado de máquina, em especial a visão computacional. Isso ocorreu principalmente por dois motivos: a disponibilidade de bases de dados com milhões de imagens~\cite{Deng09,ILSVRC15}, e de computadores capazes de reduzir o tempo necessário para realizar o processamento dessas bases de dados. Antes disso, alguns estudos exploraram a utilização de representações hierárquicas com redes neurais, tais como o Neocognitron de Fukushima~\cite{Fukushima88} e a rede neural para reconhecimento de dígitos de LeCun~\cite{Lecun98}. Apesar desses métodos serem conhecidos pelas comunidades de Aprendizado de Máquinas e Inteligência Artificial, grande parte dos esforços dos pesquisadores de outras áreas tiveram como foco outras técnicas. Por exemplo, em Visão Computacional abordagens baseadas em características invariantes a escala, \textit{Bag-of-Features}, pirâmides espaciais e abordagens baseadas em dicionários foram bastante utilizadas durante o final da década de 2000. Características baseadas em frequência foram comumente utilizadas pela comunidade de Processamento de Sinais, Áudio e Fala, enquanto métodos de \textit{Bag-of-Words} e suas variantes foram explorados no Processamento de Linguagem Natural.

Talvez um dos principais marcos que atraiu a atenção das diferentes comunidades tenha sido a publicação do artigo e código-fonte que implementa a rede neural convolucional AlexNet~\cite{Krizhevsky12}. Os resultados apontavam para uma indiscutível performance estatística de tais métodos para realizar classificação de imagem e, a partir desse ponto, o Aprendizado Profundo (do inglês \textit{Deep Learning} (DL)) passou a ser aplicado a diversas áreas, em particular, mas não exclusivamente, as áreas de Visão Computacional, Processamento de Imagens, Computação Gráfica. Redes neurais convolucionais (do inglês \textit{Convolutional Neural Networks} (CNN)), \textit{Deep Belief Networks}, \textit{Restricted Boltzmman Machines} e \textit{Autoencoders} (AE) começaram a aparecer como base para métodos do estado da arte em diversas aplicações. A competição ImageNet~\cite{Deng09} teve grande impacto nesse processo, começando uma corrida para encontrar o modelo que seria capaz de superar o atual campeão nesse desafio de classificação de imagens, além de segmentação de imagens, reconhecimento de objetos, entre outras tarefas. 

De fato, técnicas de Aprendizado Profundo oferecem atualmente um importante conjunto de métodos para analisar sinais como áudio e fala, conteúdos visuais, incluindo imagens e vídeos, e ainda conteúdo textual. Entretanto, esses métodos incluem diversos modelos, componentes e algoritmos. A variedade de palavras-chave utilizadas nesse contexto faz com que a literatura da área praticamente se torne uma nova linguagem: Mapas de Atributos, Ativação, Campos Receptivos, \textit{dropout}, ReLU, MaxPool, softmax, SGD, Adam, FC, etc. Isso pode fazer com que seja difícil para que um leigo entenda e consiga acompanhar estudos recentes. Além disso, apesar do amplo uso de redes neurais profundas por pesquisadores em busca da resolução de problemas, há ainda uma lacuna no entendimento desses métodos: como eles funcionam, em que situações funcionam e quais suas limitações?

Nesse texto, iremos descrever as Redes Neurais Profundas (\textit{Deep Neural Networks}), e em particular as Redes Convolucionais. Como o foco é na compreensão do funcionamento desse grupo de métodos, outros métodos de Aprendizado Profundo não serão tratados tais como \textit{Deep Belief Networks} (DBN), \textit{Deep Boltzmann Machines} (DBM) e métodos que utilizam Redes Neurais Recorrentes (do inglês \textit{Recurent Neural Networks} (RNN)) e \textit{Reinforcement Learning}. Para maiores detalhes sobre esses métodos, recomenda-se ~\cite{Fischer2014training,Larochelle2012,Salakhutdinov2009deep} como referências para estudos que utilizam DBNs e DBMs, e~\cite{Zheng2015conditional,Graves2013speech,Pascanu2013difficulty} para estudos relacionados a RNNs. Para o estudo de Autoencoders, recomendamos~\cite{Bengio2013RepresentationLearningReview} e~\cite{Goodfellow16}, e para o estudo de Redes Geradoras Adversariais (Generative Adversarial Networks, GANs)~\cite{Goodfellow14}. Acreditamos que a leitura desse material será de grande ajuda para o entendimento dos outros métodos citados e não cobertos por esse texto.

A partir da seção seguinte iremos nos referir ao Aprendizado Profundo pelo seu termo em inglês \textit{Deep Learning} (DL), visto que é esse o mais comumente utilizado mesmo na literatura em lingua portuguesa. Utilizaremos ainda as siglas em inglês pelo mesmo motivo, e.g. CNN.

Esse capítulo está organizado da seguinte forma. A Seção~\ref{sec:definitions} apresenta definições que serão utilizadas ao longo do texto bem como os pré-requisitos necessários. A seguir, na Seção~\ref{sec:shallowtodeep} apresentamos de forma suave os conceitos desde o aprendizado de máquina até o aprendizado profundo, por meio do exemplo de classificação de imagens. As Redes Convolucionais (Seção~\ref{sec:cnn}) são apresentadas, contextualizando suas particularidades. Então, na Seção~\ref{sec:why} apresentamos as bases teóricas que até o momento explicam o sucesso dos métodos de aprendizado profundo. Finalmente, considerações finais são discutidas na Seção~\ref{sec:conclusions}.

\section{Deep Learning: pré-requisitos e definições }
\label{sec:definitions}

Os pré-requisitos necessários para entender como Deep Learning funciona, incluem conhecimentos básicos de Aprendizado de Máquinas (ML) e Processamento de Imagens (IP), em particular conhecimentos básicos sobre aprendizado supervisionado, classificação, redes neurais Multilayer Perceptron (MLP), aprendizado não-supervisionado, fundamentos de processamento de imagens, representação de imagens, filtragem e convolução. Como esses conceitos estão fora do escopo deste capítulo, recomenda-se~\cite{Gonzalez2007,BenDavid2014} como leituras introdutórias. Assume-se também que o leitor está familiarizado com Álgebra Linear e Cálculo, além de Probabilidade e Otimização --- uma introdução a esses tópicos pode ser encontrada na Parte 1 do livro-texto de Goodfellow~\etal sobre Deep Learning~\cite{Goodfellow16}. Ainda assim tentaremos descrever os métodos de forma clara o suficiente para que seja possível o entendimento mesmo com rudimentos dos pré-requisitos.

Métodos que utilizam \textbf{Deep Learning} buscam descobrir um modelo (e.g. regras, parâmetros) utilizando um conjunto de dados (exemplos) e um método para guiar o aprendizado do modelo a partir desses exemplos. Ao final do processo de aprendizado tem-se uma função capaz de receber por entrada os dados brutos e fornecer como saída uma representação adequada para o problema em questão. Mas como DL se diferencia de ML? Vamos ver dois exemplos.

\paragraph{Exemplo 1 --- classificação de imagens}: deseja-se receber como entrada uma imagem no formato RGB e produzir como saída as probabilidades dessa imagem pertencer a um conjunto possíveis de classes (e.g. uma função que seja capaz de diferenciar imagens que contém cães, gatos, tartarugas e corujas). Assim queremos aprender uma função como $f(\mathbf{x}) = y$, em que $x$ representa a imagem, e $y$ a classe mais provável para $\mathbf{x}$.

\paragraph{Exemplo 2 --- detecção de anomalias em sinais de fala}: deseja-se receber como entrada um áudio (sinal) e produzir como saída uma representação desse sinal que permita dizer se a entrada representa um caso normal ou anômalo (e.g. uma função que permita responder para anomalias na fala de uma pessoa que indique alguma enfermidade ou síndrome). Assim, a função a ser aprendida estaria no formato $f(\mathbf{x}) = p$, em que $p$ representa a probabilidade de $\mathbf{x}$ ser um áudio anômalo. 

Note que a definição e ambos os exemplos são iguais aos de aprendizado de máquina: o Exemplo 1 é um cenário de aprendizado supervisionado, enquanto o Exemplo 2 se trata de um cenário semi-supervisionado (podendo também ser não supervisionado a depender da base de dados). 

\paragraph{De ML para DL}: a diferença quando se trata de DL é como se aprende a função $f(.)$. De forma geral, métodos não-DL, comumente referidos na literatura como ``superficiais'' ou ``rasos'' (o termo em inglês, que utilizaremos, é \textit{shallow}) buscam diretamente por uma única função que possa, a partir de um conjunto de parâmetros, gerar o resultado desejado. Por outro lado, em DL temos métodos que aprendem $f(.)$ por meio da \textbf{composições} de funções, i.e.:
\begin{align*}
    f(\mathbf{x}) = f_L \left( \cdots f_2(f_1(\mathbf{x}_1)) \cdots) \right),
\end{align*}
onde cada função $f_l(.)$ (o índice $l$ se refere comumente a uma ``camada'', veremos mais a frente o significado desse termo) toma como entrada um vetor de dados $\mathbf{x}_l$ (preparado para a camada $l$), gerando como saída o próximo vetor $\mathbf{x}_{l+1}$.

Para sermos mais precisos, cada função faz uso de parâmetros para realizar a transformação dos dados de entrada. Iremos denotar o conjunto desses parâmetros (comumente uma matriz) por $W_l$, relacionado a cada função $f_l$, e então podemos escrever:
\begin{align*}
    f_L \left( \cdots f_2(f_1(\mathbf{x}_1, W_1);W_2) \cdots), W_L \right), 
\end{align*}
onde $\mathbf{x}_{1}$ representa os dados de entrada, cada função tem seu próprio conjunto de parâmetros e sua saída será passada para a próxima função. Na equação acima, temos a composição de $L$ funções, ou $L$ camadas.

Assim uma das ideias centrais em Deep Learning é aprender sucessivas representações dos dados, intermediárias, ou seja, os $\mathbf{x}_{l}, l=1\cdots L$ acima. Os algoritmos de DL resolvem o problema de encontrar os parâmetros $W$ diretamente a partir dos dados e definem cada representação como combinações de outras (anteriores) e mais simples~\cite{Goodfellow16}. Assim a profundidade (em termos das representações sucessivas) permite aprender uma sequência de funções que transformam vetores mapeando-os de um espaço a outro, até atingir o resultado desejado.

Por isso é de fundamental importância a hierarquia das representações: cada função opera sobre uma entrada gerando uma representação que é então passada para a próxima função. A hipótese em DL é a de que, se tivermos um número suficiente de camadas $L$, espaços com dimensionalidade alta o suficiente, i.e., o número de parâmetros $W$ em cada função (iremos detalhar o que isso significa nas próximas seções), e dados suficientes para aprender os parâmetros $W_l$ para todo $l$, então conseguiremos capturar o escopo das relações nos dados originais, encontrando assim a representação mais adequada para a tarefa desejada~\cite{Chollet2017}. Matematicamente, podemos interpretar que essa sequência de transformações separa as múltiplas variedades (do ponto de vista geométrico, do termo inglês \textit{manifolds}) que nos dados originais estariam todas enoveladas.

Na seção seguinte utilizaremos o exemplo do reconhecimento de dígitos para ilustrar os conceitos introduzidos, indo de um modelo \textit{shallow} para um modelo \textit{deep}.

\section{\textit{From Shallow to Deep}}
\label{sec:shallowtodeep}

\begin{figure}
\begin{center}
 \includegraphics[width=0.45\linewidth]{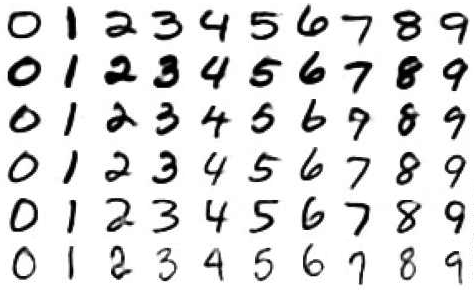}
\end{center}
 \caption{Exemplos de imagens das 10 classes da base de dígitos MNIST.}
 \label{fig:mnistdataset}
\end{figure}

% Use MNIST to go from 1-Layer to Multi-Layer (tutorial do Martin)
Vamos tomar como exemplo o problema de classificar imagens contendo dígitos numéricos dentre 10 dígitos (de 0 a 9). Para isso utilizaremos a base de dados MNIST que possui imagens de tamanho $28 \times 28$ (veja exemplos na Figura~\ref{fig:mnistdataset}). Montando uma arquitetura de rede neural simples, poderíamos tentar resolver o problema da seguinte forma:
\begin{enumerate}
    \item \textbf{entrada}: vetorizamos a imagem de entrada $28\times 28$ de forma a obter um vetor $\mathbf{x}$ de tamanho $784 \times 1$;
    \item \textbf{neurônios}: criamos 10 neurônios de saída, cada um representando a probabilidade da entrada pertencer a uma das 10 classes (dígitos 0,1,2,3,4,5,6,7,8 ou 9);
    \item \textbf{pesos/parâmetros}: assim como em uma rede MLP, cada neurônio de saída está associado a pesos e termos bias que são aplicados nos dados para gerar uma combinação linear como saída;
    \item \textbf{classificação}: para que a saída possa ser interpretada como probabilidades permitindo atribuir um exemplo à classe com maior probabilidade, utilizamos uma função conhecida como \textit{softmax}.
\end{enumerate}

Vamos recordar a montagem de uma rede neural e definir a notação. Na nossa arquitetura \textit{shallow}, cada neurônio $j$ processa uma imagem de entrada na forma $\textbf{w}_j^t\textbf{x} + b_j$. Para ilustrar, se tivermos uma entrada com 4 valores no vetor, o neurônio $j$ produziria o seguinte resultado:
\begin{equation*}
\textbf{w}_j^t\textbf{x} + b_j = [(w_{j,1} x_1) + (w_{j,2} x_2) + (w_{j,3} x_3) + (w_{j,4} x_4)] + b_j
\end{equation*}

Como temos apenas uma camada precisamos que esse valor resultante do neurônio $j$ seja o score da rede neural para a classe $j$. Assim, a partir do valor gerado pelo produto vetorial e somado o bias, aplicamos a função softmax, que é uma forma de obter valores normalizados para o intervalo 0-1 para cada classe $c$. A probabilidade de predizer $y$ como pertencendo à classe $c$, dada uma imagem de entrada $\mathbf{x}$, um conjunto de pesos, $w$, e ainda os termos \textit{bias}, $\mathbf{b}$, ambos relativos ao neurônio da classe $c$ é definida como:
\begin{equation*}
P(y=c|\mathbf{x};\mathbf{w}_c; b_c) = \operatorname{softmax}_c(\mathbf{x}^t\mathbf{w}_c+ b_c) = \frac{e^{\mathbf{x}^t\mathbf{w}_c+ b_c}}{\sum_j |e^{\mathbf{x}^t\mathbf{w}_j+ b_j}|}.
\end{equation*}

Note que então temos primeiro uma combinação linear $\mathbf{x}^t\mathbf{w}_c+b_c$, depois exponenciamos esse valor, e dividimos pela soma da saída de todos os outros neurônios, de forma que a soma seja unitária, respeitando os critérios para uma distribuição de probabilidade. A função softmax é considerada uma \textbf{função de ativação} para classificação. Veremos posteriormente que existem diversas funções de ativação possíveis: cada qual adequada a um tipo de cenário.

\begin{figure}[!ht]
\begin{center}
\begin{tabular}{cc}
    \begin{tikzpicture}[node distance=0.0cm,scale=1.25, every node/.style={scale=0.75}]
        \node[draw,minimum size=0.6cm, name=x1, fill=white] (x1) {$x_1$};
    
        \node[draw,minimum size=0.6cm, right=of x1, xshift=3.5cm, yshift=4.0cm] (n1) {};
    	\foreach \i [count=\j] in {2,...,10}
            \node[draw,minimum size=0.6cm, below=of n\j, yshift=-0.2cm] (n\i) {};

    	\foreach \i [count=\j] in {1,...,10}
            \draw [->,gray,-latex] (n\i) -- ++(0.6,0) node [right] {$\operatorname{softmax}_{\j}(w_{\j}x_1+b_{\j})$};
    
    	\foreach \i in {1}
      	   \foreach \j in {1,...,10}
    	      \draw [->,gray,-latex] (x\i) -- (n\j.west); % node [above,pos=0.85,black] {$w_{1,\j}$};

        % \draw[->,line width=0.5pt] (input.east) -- (encoder.west);
        % \draw[->,line width=0.5pt] (encoder.east) -- (code.west);
        % \draw[->,line width=0.5pt] (code.east) -- (decoder.west);
        % \draw[->,line width=0.5pt] (decoder.east) -- (output.west);
    \end{tikzpicture}
&
    \begin{tikzpicture}[node distance=0.0cm,scale=1.25, every node/.style={scale=0.75}]
        \node[draw,minimum size=0.6cm, name=x1, fill=black] (x1) {};
        \node[draw,minimum size=0.6cm, below=of x1, fill=black] (x2) {};
        \node[draw,minimum size=0.6cm, below=of x2, fill=gray!80!white] (x3) {};
        \node[draw,minimum size=0.6cm, below=of x3, fill=white] (x4) {};
        \node[draw,minimum size=0.6cm, below=of x4, fill=gray!90!white] (x5) {};
        \node[draw,minimum size=0.6cm, below=of x5, fill=black] (x6) {};
        \node[draw,minimum size=0.6cm, below=of x6, fill=gray!90!white] (x7) {};
        \node[minimum size=0.6cm, below=of x7, fill=white] (xd) {...};
        \node[draw,minimum size=0.6cm, below=of xd, fill=black] (x8) {};
    
        \node[draw,minimum size=0.6cm, right=of x1, xshift=2.5cm, yshift=1.2cm] (n1) {};
    	\foreach \i [count=\j] in {2,...,10}
            \node[draw,minimum size=0.6cm, below=of n\j, yshift=-0.2cm] (n\i) {};
    
    	\foreach \i in {1,...,8}
      	   \foreach \j in {1,...,10}
    	      \draw [->,gray,-latex] (x\i) -- (n\j.west);
    	      
    	\foreach \i [count=\j] in {1,...,10}
            \draw [->,gray,-latex] (n\i) -- ++(0.6,0) node [right] {$\operatorname{softmax}_{\j}(\mathbf{w}_{\j}^t \mathbf{x}+b_{\j})$};

    \end{tikzpicture}

\\
(a) um único valor $x_1$ como entrada  & (b) vetor $\mathbf{x}$ como entrada

\end{tabular}
  \end{center}
  \caption{Uma arquitetura \textit{shallow}: (a) caso em que teríamos apenas um valor escalar de entrada e 10 neurônios de saída, (b) mostra um vetor de valores como entrada, assim cada um dos 10 neurônios de saída está associado a um vetor de pesos diferente.}\label{fig:shallow1}
\end{figure}
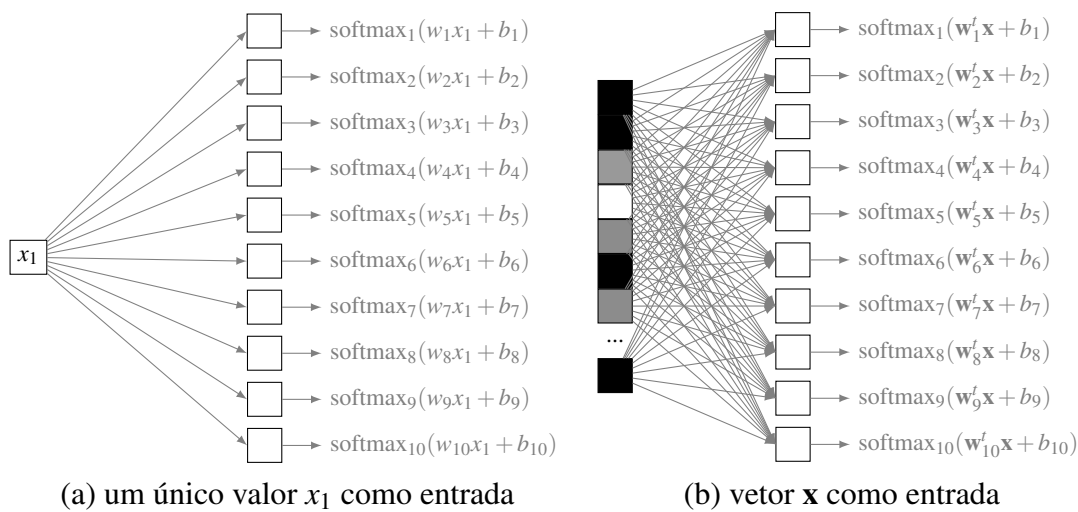

Na Figura~\ref{fig:shallow1} mostramos a arquitetura pretendida. A título de ilustração mostramos dois casos, no primeiro teríamos como entrada apenas um escalar, enquanto que, no segundo caso, tem-se um vetor como entrada.

% TODO: figura com a arquitetura planejada

Na prática, ao invés de utilizarmos uma única imagem de entrada por vez, utilizamos um conjunto de exemplos de entrada, chamado de \textbf{batch}. Para que isso possa ser feito, utilizaremos notação matricial. Agora, o resultado da combinação linear computada pelos neurônios será dado por:
\begin{equation}
\mathbf{X} \cdot \mathbf{W} + \mathbf{b}
\label{eq:matrixnn}
\end{equation}

$\mathbf{X}$ é uma matriz em que cada linha é uma imagem (ou seja, cada linha possui 784 valores, referentes aos 784 pixels da imagem). Aqui utilizaremos batches de tamanho 64, ou seja, a rede receberá como entrada 64 imagens de uma só vez e portanto a matriz $\mathbf{X}$ tem tamanho $64 \times 784$.

$\mathbf{W}$ é uma matriz que contém os pesos com os quais as imagens serão transformadas. Note que queremos produzir como saída, para cada imagem (linha da matriz $\mathbf{X}$), as probabilidades para os 10 dígitos. Para isso a matriz $\mathbf{W}$ deverá ter 10 colunas e 784 linhas. Cada coluna contém os 784 pesos de um neurônio.

$\mathbf{b}$ é um vetor de termos bias: são utilizados os mesmos biases para todas as 64 imagens no batch. Assim a Equação~\ref{eq:matrixnn} pode ser detalhada da seguinte forma: 
\begin{align*}
    \arraycolsep=1.4pt
    \begin{bmatrix}
       x_{0,0} & x_{0,1}& x_{0,2} & ... & x_{0,783} \\
       x_{1,0} & x_{0,1}& x_{1,2} & ... & x_{0,783} \\
       \vdots & \vdots&\vdots & \ddots & \vdots \\
       x_{63,0} & x_{63,1}& x_{63,2} & ... & x_{63,783}
    \end{bmatrix}
    \cdot
    \arraycolsep=1.4pt
    \begin{bmatrix}
       w_{0,0} & w_{0,1} & ... & w_{0,9} \\
       w_{1,0} & w_{1,1} & ... & w_{1,9} \\
       w_{2,0} & w_{2,1} & ... & w_{2,9} \\
       \vdots & \vdots & \ddots & \vdots \\
       w_{783,0} & w_{783,1} & ... & w_{783,9}
    \end{bmatrix}
    +
    \arraycolsep=1.4pt
    \begin{bmatrix}
       b_{0} & b_{1} & b_{2} & ... & b_{9}
    \end{bmatrix}
\end{align*}

Aplicamos então a função de ativação softmax para obter os resultados de classificação ou predições também em formato matricial $\mathbf{Y}$. Note que essa será uma matriz com 64 linhas e 10 colunas, ou seja, para cada uma das 64 imagens no batch, temos as probabilidades para as 10 classes:
\begin{align*}
    \mathbf{Y} &= \operatorname{softmax}(\mathbf{X} \cdot \mathbf{W} + \mathbf{b})\\
    \mathbf{Y} &= \arraycolsep=1.4pt
    \begin{bmatrix}
       y_{0,0} & y_{0,1} & y_{0,2} & ... & y_{0,9} \\
       y_{1,0} & y_{1,1} & y_{1,2} & ... & y_{1,9} \\
       \vdots & \vdots &\vdots & \ddots & \vdots \\
       y_{63,0} & y_{63,1} & y_{63,2} & ... & y_{63,9} 
    \end{bmatrix}
\end{align*}

Agora seria importante questionar: quando $\mathbf{Y}$ terá boas predições para as imagens de entrada, ou seja, predizendo corretamente os dígitos? Isso acontecerá apenas se os parâmetros da rede (pesos $\mathbf{W}$ e bias $\mathbf{b}$) forem adequados para o sistema. Mas como encontrar esses parâmetros e como saber quão bons esses parâmetros são?

Precisamos de uma função que compute a qualidade da predição realizada! Essa função é conhecida como \textbf{função de custo} (em inglês se utilizam os termos \textit{loss function} ou \textit{cost function}). Essa função é responsável por dizer quão longe estamos da predição ideal e portanto quantifica o ``custo'' ou ``perda'' ao aceitarmos a predição gerada pelos parâmetros atuais do modelo. Em outras palavras, qual é o custo de aceitarmos uma predição $\hat y$ sendo que a classe verdadeira é $y$? Para que possamos computar essa função precisamos nos basear em exemplos rotulados com suas classes verdadeiras, então o termo mais correto seria função de custo empírica.

Dentre as funções de custo mais utilizadas em classificação temos a \textbf{entropia cruzada} (\textit{cross-entropy}). Considerando um único exemplo cuja distribuição de probabilidade de classes real é $\mathbf{y}$ e a predição é dada por $f(\mathbf{x}) = \mathbf{\hat y}$ temos a soma das entropias cruzadas (entre a predição e a classe real) de cada classe $j$:
\begin{equation}
\ell^{(\text{ce})} = - \sum_j y_j \cdot \log(\hat y_j + \epsilon),
\label{eq:celoss}
\end{equation}
onde $\epsilon << 1$ é uma variável para evitar $\log(0)$. Vamos assumir $\epsilon = 0.0001$.

A título de exemplo sejam os seguintes vetores de distribuição de probabilidade de classes (note que ambas classe real e predita são a classe 7):
\begin{align*}
    \mathbf{y} &= \begin{bmatrix}
       0.00 & 0.00 & 0.00 & 0.00 & 0.00 &0.00 &1.00 &0.00 &0.00 &0.00 
    \end{bmatrix}\\
    \mathbf{\hat y} &= \begin{bmatrix}
       0.18 & 0.00 & 0.00 & 0.02 & 0.00 &0.00 &0.65 &0.05 &0.10 &0.00 
    \end{bmatrix}
\end{align*}
Então a entropia cruzada para esse exemplo seria:
\begin{align*}
    \ell^{(\text{ce})} = - (0 + 0 + 0 + 0 + 0 + 0 - 0.6214 + 0 + 0 + 0) = 0.6214
\end{align*}

% \begin{equation}
% \ell^{(\text{ce})}_j = - \log\left( \frac{e^{f_{y_j}}}{\sum_k e^{f_{k}}}\right).
% \label{eq:celoss}
% \end{equation}

Essa função recebe como entrada um vetor de scores, e produz valor 0 apenas no caso ideal em que $\mathbf{y} = \mathbf{\hat y}$. Em problemas de classificação, a entropia cruzada pode ser vista como a minimização da divergência de Kullback-Leibler entre duas distribuições de classes na qual a distribuição verdadeira tem entropia nula (já que possui um único valor não nulo, 1)~\cite{Ponti2017decision}. Assim, temos a minimização da log-verossimilhança negativa da classe correta, o que relaciona essa equação também com a estimação de máxima verossimilhança.

Dado um conjunto atual de parâmetros $\mathbf{W}$ e $\mathbf{b}$, o custo completo com base em um batch (ou subconjunto de treinamento) contendo $N$ instâncias é geralmente computado por meio da média dos custos de todas as instâncias $\mathbf{x}_i$ e suas respectivas distribuições de classe: $\mathbf{y}_i$:
\begin{equation*}
\mathcal{L}(\mathbf{W};\mathbf{b}) = \frac{1}{N}\sum_{i=1}^{N} \ell\left( \mathbf{y}_i, f(\mathbf{x}_i; \mathbf{W}; \mathbf{b})\right).
\end{equation*}
Agora, precisamos ajustar os parâmetros de forma a minimizar $\mathcal{L}(\mathbf{W};\mathbf{b})$. Isso porque comumente esses parâmetros são inicializados com valores aleatórios, e queremos modificar esses valores iniciais de maneira a \textit{convergir} para um modelo que nos dê boas predições.

Esse processo é feito utilizando algoritmos de otimização como o Gradiente Descendente (GD), que basicamente computa derivadas parciais de forma a encontrar, para cada parâmetro do modelo, qual modificação dos parâmetros permite minimizar a função. Veremos nas seções seguintes os métodos de otimização comumente utilizados em Deep Learning. Por enquanto assumiremos que o leitor está familiarizado com o Gradiente Descendente (GD) básico. Assim, executamos o treinamento por meio do algoritmo conhecido por Backpropagation, que irá atualizar os parâmetros de forma que a saída $\mathbf{\hat y}$ se aproxime do resultado esperado $\mathbf{y}$.

\paragraph{Treinamento} nesse ponto, temos todos os componentes necessários para executar o algoritmo de treinamento. Inicializamos os parâmetros de forma aleatória, e então o algoritmo: (1) carrega 64 imagens sorteadas aleatoriamente do conjunto de treinamento, e (2) ajusta os parâmetros do modelo utilizando essas 64 imagens. Os passos (1) e (2) são repetidos por um número de vezes (iterações), até que o erro de classificação computado dentro do conjunto de treinamento seja suficientemente baixo, ou estabilize. Por exemplo, podemos definir 1000 iterações. Note que, nesse caso, o treinamento utilizará no máximo $64 \times 1000 = 64000$ exemplos (imagens) para aprender os parâmetros da rede.

Para mostrar como isso poderia ser feito na prática, abaixo mostramos um código na Listagem~\ref{code:mnist1} em linguagem \texttt{Python} utilizando a biblioteca \texttt{Tensorflow} versão 1.2. Definimos primeiramente as variáveis: na linha 3 a matriz de imagens com $28\times 28\times 1$ (pois são em escala de cinza --- para imagens RGB usar 3) além de um campo para indexar as imagens no batch (indicado por \texttt{None}); na linha 4 a matriz de pesos e na linha 5 o vetor de bias. O modelo considera as imagens redimensionadas para um vetor com 784 elementos (ver linha 10). A seguir a função de custo e método de otimização são definidos e 1000 iterações são executadas. Note que esse capítulo não pretende ser um tutorial sobre Tensorflow: o código é mostrado em alguns pontos para auxiliar no entendimento dos conceitos via implementação dos métodos.

\noindent \begin{minipage}{\linewidth}
\begin{lstlisting}[caption=Treinamento de uma rede shallow com Tensorflow, label=code:mnist1, frame=single]
import tensorflow as tf
tf.GraphKeys.VARIABLES = tf.GraphKeys.GLOBAL_VARIABLES
from tensorflow.examples.tutorials.mnist import input_data as mnist_data
# variaveis (matrizes e vetores) 
X = tf.placeholder(tf.float32, [None, 28, 28, 1]) # batch de imagens X
W = tf.Variable(tf.zeros([784, 10])) # pesos
b = tf.Variable(tf.zeros([10])) # bias
Y = tf.placeholder(tf.float32, [None, 10]) # classes das imagens em X
inicia = tf.global_variables_initializer() # instancia inicializacao

# modelo que ira gerar as predicoes com base nas imagens vetorizadas
Y_ = tf.nn.softmax(tf.matmul(tf.reshape(X, [-1, 784]), W) + b)

# define funcao de custo (entropia cruzada)
entropia_cruzada = -tf.reduce_sum(Y * tf.log(Y_+0.0001))

# otimizacao com taxa de aprendizado 0.0025
otimiza = tf.train.GradientDescentOptimizer(0.0025)
treinamento = otimiza.minimize(entropia_cruzada)

sess = tf.Session() # instancia sessao
sess.run(inicia)    # executa sessao e inicializa
# baixa base de dados mnist
mnist = mnist_data.read_data_sets("data", one_hot=True, reshape=False, validation_size=0)

# executa 1000 iteracoes
for i in range(1000):
    # carrega batch de 64 imagens (X) e suas classes (Y)
    batch_X, batch_Y = mnist.train.next_batch(64)
    dados_trein={X: batch_X, Y: batch_Y}

    # treina com o batch atual
    sess.run(treinamento, feed_dict=dados_trein)
    # computa entropia-cruzada para acompanhar convergencia
    ce = sess.run(entropia_cruzada, feed_dict=dados_trein)
\end{lstlisting}
\end{minipage}

\subsection{Criando uma rede profunda} 

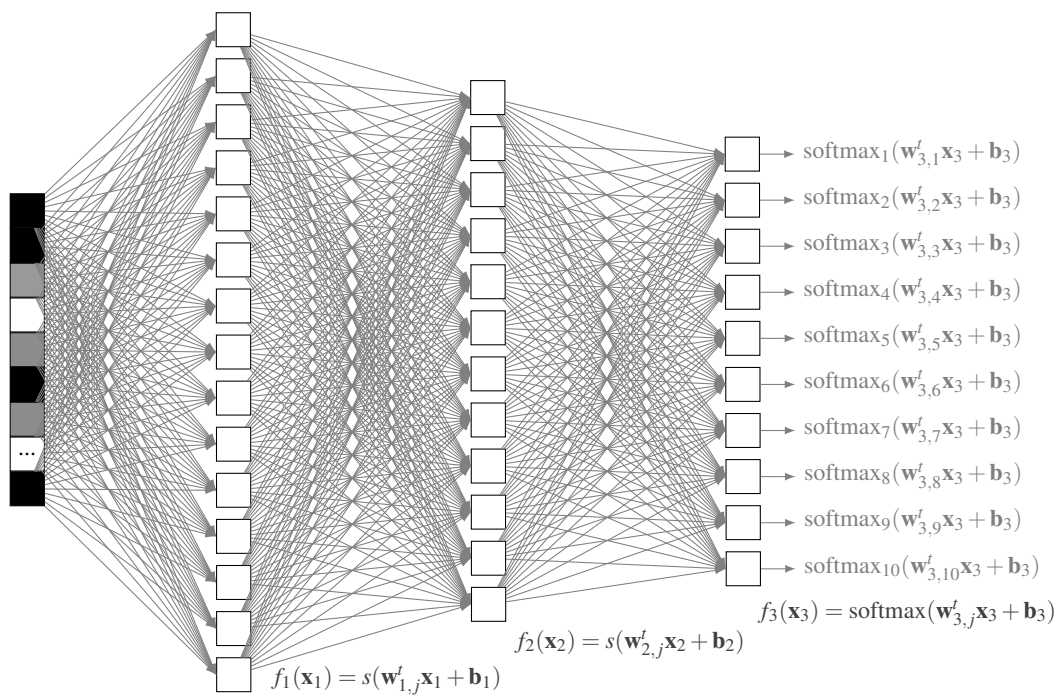
\begin{figure}[!ht]
\begin{center}
    \begin{tikzpicture}[node distance=0.0cm,scale=1.15, every node/.style={scale=0.75}]

    \node[draw,minimum size=0.6cm, name=x1, fill=black] (x1) {};
    \node[draw,minimum size=0.6cm, below=of x1, fill=black] (x2) {};
    \node[draw,minimum size=0.6cm, below=of x2, fill=gray!80!white] (x3) {};
    \node[draw,minimum size=0.6cm, below=of x3, fill=white] (x4) {};
    \node[draw,minimum size=0.6cm, below=of x4, fill=gray!90!white] (x5) {};
    \node[draw,minimum size=0.6cm, below=of x5, fill=black] (x6) {};
    \node[draw,minimum size=0.6cm, below=of x6, fill=gray!90!white] (x7) {};
    \node[draw,minimum size=0.6cm, below=of x7, fill=white] (x8) {...};
    \node[draw,minimum size=0.6cm, below=of x8, fill=black] (x9) {};

        \node[draw,minimum size=0.6cm, right=of x1, xshift=3.0cm, yshift=3.2cm] (h1) {};
    	\foreach \i [count=\j] in {2,...,15}
            \node[draw,minimum size=0.6cm, below=of h\j, yshift=-0.2cm] (h\i) {};
            
        \node[minimum size=0.6cm, right=of h15,darkgray,xshift=0.2cm,yshift=-0.1cm] (xs2) {$f_1(\mathbf{x}_1) = s(\mathbf{w}_{1,j}^{t} \mathbf{x}_1+\mathbf{b}_{1})$};

        \node[draw,minimum size=0.6cm, right=of h1, xshift=3.85cm, yshift=-1.2cm] (hh1) {};
    	\foreach \i [count=\j] in {2,...,12}
            \node[draw,minimum size=0.6cm, below=of hh\j, yshift=-0.2cm] (hh\i) {};
        
        \node[minimum size=0.6cm, right=of hh12,darkgray,xshift=0.0cm,yshift=-0.7cm] (xs3) {$f_2(\mathbf{x}_2) = s(\mathbf{w}_{2,j}^{t} \mathbf{x}_2+\mathbf{b}_{2})$};

        \node[draw,minimum size=0.6cm, right=of hh1, xshift=3.85cm, yshift=-1cm] (n1) {};
    	\foreach \i [count=\j] in {2,...,10}
            \node[draw,minimum size=0.6cm, below=of n\j, yshift=-0.2cm] (n\i) {};
    
    	\foreach \i in {1,...,9}
      	   \foreach \j in {1,...,15}
    	      \draw [->,gray,-latex] (x\i) -- (h\j.west);

    	\foreach \i in {1,...,15}
      	   \foreach \j in {1,...,12}
    	      \draw [->,gray,-latex] (h\i) -- (hh\j.west);

    	\foreach \i in {1,...,12}
      	   \foreach \j in {1,...,10}
    	      \draw [->,gray,-latex] (hh\i) -- (n\j.west);
    	      
    	\foreach \i [count=\j] in {1,...,10}
            \draw [->,gray,-latex] (n\i) -- ++(0.6,0) node [right] {$\operatorname{softmax}_{\j}(\mathbf{w}_{3,\j}^t \mathbf{x}_3+\mathbf{b}_{3})$};

        \node[minimum size=0.6cm, right=of n10, darkgray,xshift=-0.2cm,yshift=-0.8cm] (ns3) {$f_3(\mathbf{x}_3) = \operatorname{softmax}(\mathbf{w}_{3,j}^{t} \mathbf{x}_3+\mathbf{b}_{3})$};

    \end{tikzpicture}
  \end{center}
  \caption{Uma arquitetura profunda com duas camadas ocultas, gerando representações intermediárias que antecedem a classificação. A quantidade de neurônios em cada camada é meramente ilustrativa.}\label{fig:deep1}
\end{figure}

A rede anterior consegue alcançar uma acurácia próxima a $91\%$ considerando o conjunto de testes da base de dados MNIST. Para que possamos melhorar esse resultado iremos utilizar uma arquitetura profunda, fazendo uso da composição de funções. Adicionaremos 2 camadas novas entre a entrada e a saída. Essas camadas são conhecidas como camadas ocultas (\textit{hidden layers}). Em particular faremos uso de camadas completamente conectadas ou como se diz em terminologia Deep Learning, \textit{fully connected} (FC), que é nada mais do que uma camada oculta assim como utilizada em redes MLP. Ilustramos a arquitetura na Figura~\ref{fig:deep1}. Teremos portanto uma predição alcançada por uma composição na forma:
\begin{align*}
    \mathbf{\hat y} = f(\mathbf{x}) = f_3 \left( f_2(f_1(\mathbf{x}_1; \mathbf{W}_1; \mathbf{b}_1);\mathbf{W}_2; \mathbf{b}_2)), \mathbf{W}_3; \mathbf{b}_3 \right),
\end{align*}
em que $f_1(\mathbf{x}_1) = \mathbf{x}_2$, $f_2(\mathbf{x}_2) = \mathbf{x}_3$ e finalmente $f_3(\mathbf{x}_3) = \mathbf{y}$.

\begin{figure}
\begin{center}
\begin{tabular}{cc}
    \begin{tikzpicture}
	  \tikzset{plot/.style = {line width=2pt}}
	  \draw[scale=1.0,->,gray] (-2.0,0) -- (2.0,0) node[right] {$x$};
	  \draw[scale=1.0,dashed,gray] (-2.0,1) -- (2.0,1) node[above,pos=0.4] {1};
	  \draw[scale=1.0,dashed,gray] (-2.0,-1) -- (2.0,-1) node[above,pos=0.4] {-1};
	  \draw[scale=1.0,->,gray] (0,-1.2) -- (0,1.4) node[above] {$tanh(x)$};
	  \draw[scale=1.0,domain=-2:2,smooth, line width=1.5pt, variable=\x,red!70!black] plot ({\x},{tanh(\x*2)});
	\end{tikzpicture} &
	   \begin{tikzpicture}
	  \tikzset{plot/.style = {line width=2pt}}
	  \draw[scale=1.0,->,gray] (-2.0,0) -- (2.0,0) node[right] {$x$};
	  \draw[scale=1.0,dashed,gray] (-2.0,2) -- (2.0,2) node[above,pos=0.4] {1};
	  %\node[gray] (-1.5,1.5) {0};
	  \draw[scale=1.0,->,gray] (0,-0.2) -- (0,2.4) node[above] {$logistic(x)$};
	  \draw[scale=1.0,domain=-2.0:2.0,smooth, line width=1.5pt, variable=\x,red!70!black] plot ({\x},{1/(1+exp(-\x*4))*2});
	\end{tikzpicture}
	\\ 
     (a) tangente hiperbólica & (b) logistica \\
	\begin{tikzpicture}
	  \tikzset{plot/.style = {line width=2pt}}
	  \draw[scale=1.0,->,gray] (-2.0,0) -- (2.0,0) node[right] {$x$};
	  \draw[scale=1.0,->,gray] (0,-1.2) -- (0,1.4) node[above] {$max[0,x]$};
	  \draw[scale=1.0,dashed,gray] (-2.0,1) -- (2.0,1) node[above,pos=0.4] {1};
	  \draw[scale=1.0,dashed,gray] (-2.0,-1) -- (2.0,-1) node[above,pos=0.4] {-1};
	  \draw[scale=1.0,domain=-2.0:0,smooth, line width=1.5pt, variable=\x,red!70!black] plot ({\x},{\x*0)});
	  \draw[scale=1.0,domain=0:1.4,smooth, line width=1.5pt, variable=\x,red!70!black] plot ({\x},{\x});
	\end{tikzpicture} &
    \begin{tikzpicture}
	  \tikzset{plot/.style = {line width=2pt}}
	  \draw[scale=1.0,->,gray] (-2.0,0) -- (2.0,0) node[right] {$x$};
	  \draw[scale=1.0,->,gray] (0,-1.05) -- (0,1.4) node[above] {$max[ax,x], a = 0.1$};
	  \draw[scale=1.0,dashed,gray] (-2.0,1) -- (2.0,1) node[above,pos=0.4] {1};
	  \draw[scale=1.0,dashed,gray] (-2.0,-1) -- (2.0,-1) node[above,pos=0.4] {-1};
	  \draw[scale=1.0,domain=-2.0:0,smooth, line width=1.5pt, variable=\x,red!70!black] plot ({\x},{\x*0.1)});
	  \draw[scale=1.0,domain=0:1.4,smooth, line width=1.5pt, variable=\x,red!70!black] plot ({\x},{\x});
	\end{tikzpicture} 
	\\(c) ReLU & (d) PReLU
\end{tabular}
\end{center}
 \caption{Ilustração comparativa de funções de ativação: (c) ReLU é a mais utilizada em Deep Learning.}
 \label{fig:activationfunctions}
\end{figure}
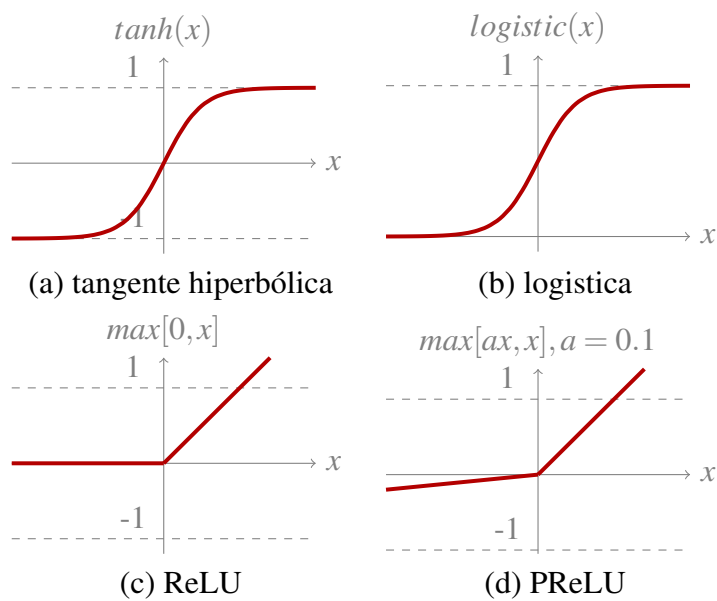

\paragraph{Funções de ativação} para as camadas FC ocultas a função de ativação softmax não é ideal! Em redes neurais é comum o uso de funções sigmoidais (como a logística e a tangente hiperbólica). Porém em Deep Learning, a função retificadora linear (\textit{rectified linear function}, ReLU) tem sido mais utilizada por facilitar o processo de treinamento~\cite{Nair10}. Isso porque as funções sigmoidais saturam a partir de um determinado ponto, enquanto a ReLU é simplesmente a função identidade para valores positivos. Veja a Figura~\ref{fig:activationfunctions} para exemplos de funções de ativação: as funções sigmoidais comprimem a saída para um intervalo curto, enquanto ReLU cancela todos os valores negativos, sendo linear para os positivos. 

A função ReLU possui relações com a restrição de não-negatividade presente em regularização como em restauração de imagens utilizando projeções em subespaços~\cite{Ponti2016imagerestoration}. Note ainda que ao calcularmos a derivada da ReLU, seu gradiente terá sempre uma direção não-nula, enquanto no caso das sigmoidais, para valores longe da origem podemos ter gradiente nulo. A ReLU paramétrica (PReLU) é uma variação que permite valores negativos com menor ponderação, parametrizado por uma variável $0\leq a\leq1$~\cite{He2015delving}. Uma das possíveis vantagens da PReLU é a possibilidade de aprender $a$ durante a fase de treinamento. No caso particular em que temos um valor fixo $a=0.01$, temos a função conhecida por \textit{Leaky ReLU}.

Na Listagem~\ref{code:mnist2} mostramos como modificar o código anterior de forma a criar camadas intermediárias e utilizar a função ReLU para essas camadas.

\noindent \begin{minipage}{\linewidth}
\begin{lstlisting}[caption=Treinamento de uma rede deep com Tensorflow, label=code:mnist2, frame=single]
# cria e inicializa aleatoriamente os pesos com distribuicao normal e sigma=0.1
W1 = tf.Variable(tf.truncated_normal([784, 64], stddev=0.1))
# bias sao inicializados com valores fixos 1/10
B1 = tf.Variable(tf.ones([64])/10)
W2 = tf.Variable(tf.truncated_normal([64, 32], stddev=0.1))
B2 = tf.Variable(tf.ones([32])/10)
W3 = tf.Variable(tf.truncated_normal([32, 10], stddev=0.1))
B3 = tf.Variable(tf.zeros([10]))

# entrada redimensionada
X1 = tf.reshape(X, [-1, 784])

# modelos das representacoes intermediarias
X2 = tf.nn.relu(tf.matmul(X1, W1) + B1)
X3 = tf.nn.relu(tf.matmul(X2, W2) + B2)

# saida da rede (a.k.a. logits)
X4 = tf.matmul(Y3, W3) + B3
# classificacao softmax
Y_ = tf.nn.softmax(X4)

# utilizamos uma funcao pronta no TF para calculo da entropia cruzada
entr_cruz = tf.nn.softmax_cross_entropy_with_logits(logits=X4, labels=Y)
entr_cruz = tf.reduce_mean(entr_cruz)*100
\end{lstlisting}
\end{minipage}

Basicamente adicionamos novas matrizes de pesos e bias. Sendo que o números de neurônio de uma camada será o tamanho do seu vetor de saída. Assim, a matriz de pesos da camada seguinte deverá respeitar esse número. No código mostramos as camadas ocultas com 64 e 32 neurônios. Portanto $\mathbf{W}_1$ terá tamanho $784\times 64$ (lembre que 784 é o tamanho do vetor de entrada), $\mathbf{W}_2$ terá tamanho $64\times 32$ e finalmente $\mathbf{W}_3$ terá tamanho $32\times 10$. Note também no código que a inicialização das matrizes de peso é feita usando números aleatórios obtidos de uma distribuição normal. Os bias são inicializados com valores pequenos, no exemplo todos iguais a 1/10.

\paragraph{Taxa de aprendizado com decaimento}: é comum definir a taxa de aprendizado com decaimento (ao invés de fixa como no exemplo anterior). Isso significa que podemos começar com um valor mais alto, por exemplo algo entre $0.005$ e $0.0025$, e definir uma função de decaimento exponencial como $\exp^{-k/d}$, em que $k$ é a iteração atual e $d$ a taxa de decaimento (quanto maior, mais lento o decaimento). 

Utilizando a base MNIST, com uma arquitetura parecida com a definida acima, é possível alcançar acurácias próximas a $97\%$ no conjunto de testes. Isso foi alcançado por meio da inclusão de novas camadas, o que torna mais fácil encontrar os parâmetros corretos pois agora não é mais necessário aprender uma transformação direta de um vetor para uma classe, mas ao invés, aprendemos representações intermediárias, da mais simples, que processa o vetor, até a mais complexa, que prediz a classe da imagem. Sendo a MNIST uma base de imagens, para ir além do resultado atual precisamos utilizar um tipo de rede conhecida como Rede Convolucional.

\section{Redes Convolucionais (CNNs)}
\label{sec:cnn}
%TODO: MP
%Basic textbooks: Goodfellow et al~\cite{Goodfellow16}, Chollet~\cite{Chollet2017}; Nielsen~\cite{Nielsen17}.
Redes Neurais Convolucionais (CNNs) são provavelmente o modelo de rede Deep Learning mais conhecido e utilizado atualmente. O que caracteriza esse tipo de rede é ser composta basicamente de \textbf{camadas convolucionais}, que processa as entradas considerando campos receptivos locais. Adicionalmente inclui operações conhecidas como \textit{pooling}, responsáveis por reduzir a dimensionalidade espacial das representações. Atualmente as CNNs de maior destaque incluem as Redes Residuais (ResNet)~\cite{He15} e Inception~\cite{Szegedy2016rethinking}. 

A principal aplicação das CNNs é para o processamento de informações visuais, em particular imagens, pois a convolução permite filtrar as imagens considerando sua estrutura bidimensional (espacial). Considere o exemplo da base de dados MNIST apresentado na seção anterior. Note que ao vetorizar as imagens estamos desprezando toda a estrutura espacial que permite entender a relação entre os pixels vizinhos em uma determinada imagem. É esse tipo de estrutura que as CNNs tentam capturar por meio da camada convolucional.

\subsection{Camada convolucional}

Na camada convolucional cada neurônio é um filtro aplicado a uma imagem de entrada e cada filtro é uma matriz de pesos. Novamente, indicamos o livro texto~\cite{Gonzalez2007} para uma introdução sobre convolução e filtros de imagens.

Seja uma imagem RGB de tamanho $224 \times 224 \times 3$ (o 3 indica os canais de cor R, G e B), que serve de entrada para uma camada convolucional. Cada filtro (neurônio) dessa camada irá processar a imagem e produzir uma transformação dessa imagem por meio de uma combinação linear dos pixels vizinhos. Note que agora não temos um peso para cada elemento da imagem. A título de exemplo em uma camada FC teríamos, para cada neurônio do nosso exemplo, 150528 pesos, um para cada valor de entrada. Ao invés, definimos filtros de tamanho $k\times k \times d$, em que $k$ é a dimensão espacial do filtro (a ser definida) e $d$ a dimensão de profundidade (essa depende da entrada da camada). Por exemplo, se definirmos $k=5$ para a a primeira camada convolucional, então teremos filtros $5\times 5 \times 3$, pois como a imagem possui 3 canais (RGB), então $d=3$, e assim cada neurônio terá $5\times 5 \times 3 = 75$ pesos.

\begin{figure}
\begin{center}
 \includegraphics[width=0.55\linewidth]{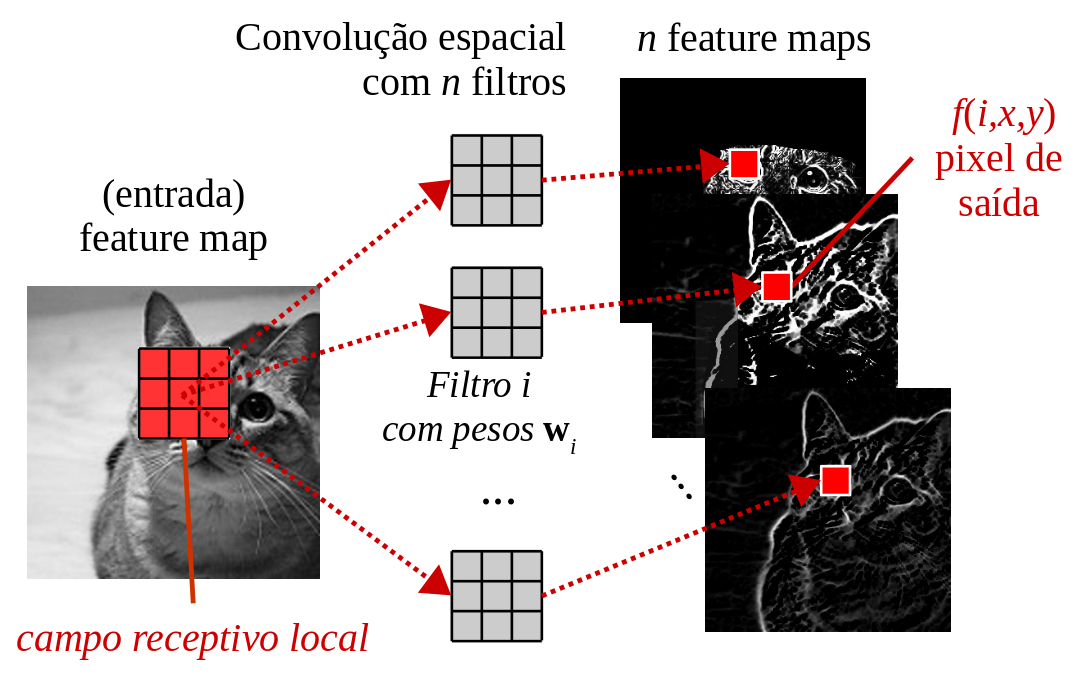}
\end{center}
 \caption{Ao utilizar convolução, processa-se informações locais utilizando cada posição $(x,y)$ como centro: essa região é chamada de campo receptivo. Seus valores são então usados como entrada para um filtro $i$ com parâmetros $\mathbf{w}_i$, produzindo um único valor (pixel) no mapa de características $f(i,x,y)$ gerado como saída.}
 \label{fig:CNNlrf}
\end{figure}

Cada região da imagem processada pelo filtro é chamada de campo receptivo local (\textit{local receptive field}); um valor de saída (pixel) é uma combinação dos pixels de entrada nesse campo receptivo local (veja Figura~\ref{fig:CNNlrf}). No entanto todos os campos receptivos são filtrados com os mesmos pesos locais para todo pixel. Isso é o que torna a camada convolucional diferente da camada FC. Assim, um valor de saída ainda terá o formato bidimensional. Por exemplo podemos dizer que $f_{l+1}(i,x,y)$ é o pixel resultante da filtragem da imagem vinda da camada anterior $l$, processada pelo filtro $i$ a partir dos valores da vizinhança centrados na posição $(x,y)$. No nosso exemplo com $k=5$, teremos uma combinação linear de 25 pixels da vizinhança para gerar um único pixel de saída.

Os tamanhos de filtros mais utilizados são $5\times 5 \times d$, $3\times 3 \times d$ e $1\times 1 \times d$. Como trabalhamos com matrizes multidimensionais (com profundidade $d$), utilizamos o termo \textbf{tensor} para denotá-las. 

Considere um problema em que temos como entrada imagens RGB, de tamanho $64 \times 64 \times 3$. Sejam então duas camadas convolucionais, a primeira com 4 filtros de tamanho $k_1= 5$, e a segunda com 5 filtros de tamanho $k_2=3$. Considere ainda que a convolução é feita utilizando extensão da imagem com preenchimento por zeros (\textit{zero padding}) de forma que conseguimos realizar a filtragem para todos os pixels da imagem, mantendo seu tamanho. Nesse cenário teríamos a seguinte composição:
\begin{align*}
    \mathbf{\hat y} = f(\mathbf{x}) = f_2(f_1(\mathbf{x}_1; \mathbf{W}_1; \mathbf{b}_1);\mathbf{W}_2; \mathbf{b}_2)),
\end{align*}
em que $\mathbf{W}_1$ possui dimensão $4 \times 5 \times 5 \times 3$ (4 filtros de tamanho $5 \times 5$, entrada com profundidade 3), e portanto a saída da camada 1, $\mathbf{x}_2 = f_1(\mathbf{x}_1)$ terá tamanho: $64\times 64 \times 4$. Após a convolução utiliza-se uma função de ativação (comumente a função ReLU já descrita anteriormente) que trunca para zero os pixels negativos. A Figura~\ref{fig:CNNfeaturemaps} ilustra esse processo, bem como o da camada 2, que recebe por entrada o tensor $64\times 64 \times 4$. Essa segunda camada possui 5 filtros $3\times 3 \times 4$ (já que a profundidade do tensor de entrada tem $d=4$), e gera como saída $\mathbf{x}_3 = f_2(\mathbf{x}_2)$, um tensor de tamanho $64\times 64 \times 5$. 

Chamamos de mapa de características a saída de cada neurônio da camada convolucional (mais detalhes na seção seguinte). Antes de prosseguir, outro aspecto importante para se mencionar é o passo ou \textit{stride}. A convolução convencionais é feita com passo/stride 1, ou seja, filtramos todos os pixels e portanto para uma imagem de entrada de tamanho $64\times 64$, geramos uma nova imagem de tamanho $64\times 64$. O uso de strides maiores que 1 é comum quando deseja-se reduzir o tempo de execução, pulando pixels e assim gerando imagens menores. Ex. com stride = 2 teremos como saída uma imagem de tamanho $32\times 32$.

\subsection{Feature maps (mapas de características)} 

\begin{figure*}
\begin{center}
 \includegraphics[width=0.95\linewidth]{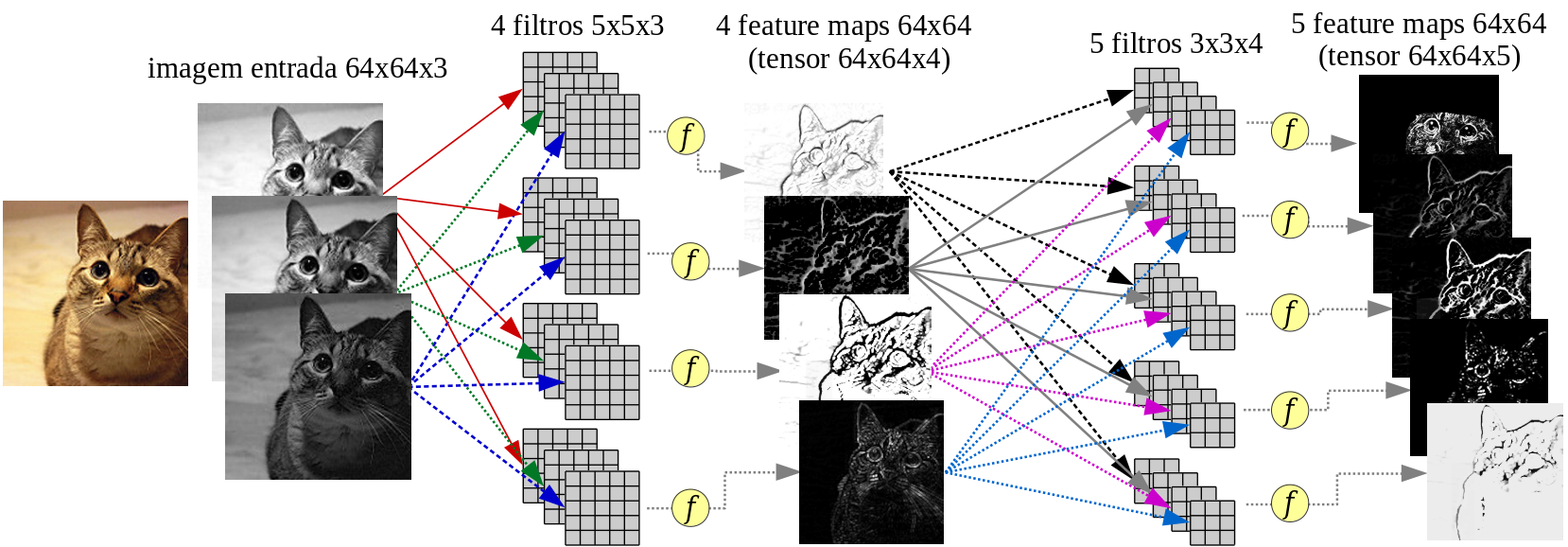}
\end{center}
 \caption{Ilustração de duas camadas convolucionais: a primeira com 4 filtros $5\times 5 \times 3$, que recebe como entrada uma imagem RGB $64\times 64\times 3$, e produz um tensor com 4 feature maps; a segunda camada convolucional contém 5 filtros  $3\times 3 \times 4$ que filtram o tensor da camada anterior, produzindo um novo tensor de feature maps com tamanho $64 \times 64 \times 5$. Os círculos após cada filtro denotam funções de ativação, como por exemplo a ReLU.}
 \label{fig:CNNfeaturemaps}
\end{figure*}

Cada representação gerada por um filtro da camada convolucional é conhecida como ``mapa de características'', do inglês \textit{feature map} ou \textit{activation map}. Utilizaremos \textit{feature map} pois é o termo mais comum na literatura. Os mapas gerados pelos diversos filtros da camada convolucional são empilhados, formando um tensor cuja profundidade é igual ao número de filtros. Esse tensor será oferecido como entrada para a próxima camada como mostrado na Figura~\ref{fig:CNNfeaturemaps}. Note que, como a primeira camada convolucional gera um tensor $64\times 64\times 4$, os filtros da segunda camada terão que ter profundidade $4$. Caso adicionássemos uma terceira camada convolucional, os filtros teriam que ter profundidade $5$.

\subsection{Pooling} 
É comum reduzir a dimensão espacial dos mapas ao longo das camadas da rede. Essa redução em tamanho é chamada de \textit{pooling} sendo a operação de máximo \textit{maxpooling} comumente empregada. Essa operação tem dois propósitos: em primeiro lugar, o custo computacional, pois como a profundidade dos tensores, $d$, costuma aumentar ao longo das camadas, é conveniente reduzir a dimensão espacial dos mesmos. Em segundo, reduzindo o tamanho das imagens obtemos um tipo de composição de banco de filtros multi-resolução que processa imagens em diferentes espaços-escala. Há estudos a favor de não utilizar pooling, mas aumentar o stride nas convoluções~\cite{Springenberg2015striving}, o que produz o mesmo efeito redutor. 

\subsection{Camadas fully connected (FC)}

Já mencionamos o uso de camadas FC, que são camadas presentes em redes neurais MLP. Nesse tipo de camada cada neurônio possui um peso associado a cada elemento do vetor de entrada. Em CNNs utiliza-se camadas FC posicionadas após múltiplas camadas convolucionais. A transição entre uma camada convolucional (que produz um tensor) e uma camada FC, exige que o tensor seja vetorizado. Por exemplo, se a camada convolucional antes de uma camada FC gera um tensor $4\times 4\times 40$, redimensionamos esses dados de forma que ele possua tamanho $1\times (4\times 4\times 40) = 1 \times 640$. Assim, cada neurônio na camada FC deverá possuir $640$ pesos de forma a produzir uma combinação linear do vetor.  

Como descrito anteriormente, arquiteturas mais recentes utilizam camadas FC ocultas com função de ativação ReLU, e a camada de saída (classificador) com função de ativação softmax.

\subsection{Arquiteturas de CNNs e seus parâmetros}

CNNs tradicionais são combinação de blocos de camadas convolucionais (Conv) seguidas por funções de ativação, eventualmente utilizando também \textit{pooling} (Pool) e então uma séria de camadas completamente conectadas (FC), também acompanhadas por funções de ativação, da seguinte forma: 
\begin{equation*}
\operatorname{CNN} \equiv P \times \left[ C\times \left( \text{Conv} \rightarrow \text{AF} \right) \rightarrow \text{Pool} \right] \rightarrow F \times \left[ \text{FC} \rightarrow \text{AF} \right]
\end{equation*}

Para criar uma CNN, deve-se definir os seguintes hiper-parâmetros: número de camadas convolucionais $C$ (para cada camada, o número de filtros, seu tamanho e o tamanho do passo dado durante a convolução), número de camadas de \texttt{pooling} $P$ (sendo nesse caso necessário escolher também o tamanho da janela e do passo que definirão o fator de subamostragem), o número de camadas totalmente conectadas $F$ (e o número de neurônios contidos em cada uma dessas camadas).

Note no entanto que algumas redes mais recentes também tem empregado a pré-ativação (a função AF é aplicada nos dados de entrada, antes da convolução ou camada FC).

O número de parâmetros em uma CNN está relacionado basicamente, aos valores a serem aprendidos em todos todos os filtros nas camadas convolucionais, os pesos das camadas totalmente conectadas e os bias.

\noindent---\textit{Exemplo}: considere uma arquitetura para analisar imagens RGB com dimensão $64\times 64 \times 3$ cujo objetivo é classificar essas imagens em 5 classes. Essa arquitetura será composta por três camadas convolucionais, duas \textit{max pooling}, e duas camadas totalmente conectadas, da seguinte forma:
\begin{itemize}
    \item Conv.L 1: 10 filtros $5\times 5 \times 3$, saída: tensor de dimensão $64\times 64\times 10$ 
    
    \item \textit{Max pooling} 1: subamostragem com fator 4 (janela de tamanho $2\times 2$ e stride 2), \\
    saída: tensor de dimensão $16\times 16\times 10$
    
    \item Conv.L2: 20 filtros $3\times 3 \times 10$, saída: tensor de dimensão $16\times 16\times 20$
    
    \item Conv.L3: 40 filtros $1\times 1 \times 20$, saída: tensor de dimensão $16\times 16\times 40$
    
    \item \textit{Max pooling} 2: subamostragem com fator 4 (janela de tamanho $2\times 2$ e stride 2),\\
    saída: tensor de dimensão $4\times 4\times 40$
    
    \item FC.L1: 32 neurônios, saída: $32$ valores
    
    \item FC.L2 (saída da rede): 5 neurônios (um por classe), saída: $5$ valores
\end{itemize}

Considerando que cada um dos filtros das três camadas convolucionais tem $p \times q \times d$ parâmetros, além do bias, e que as camadas totalmente conectadas possuem pesos e o termo bias associado com cada valor do vetor recebido da camada anterior, então o número total de parâmetros nessa arquitetura é:
\begin{align*}
     ~& (10 \times [5\times 5 \times 3 + 1] = 760) &~\text{[Conv.L1]}\\
    +~& (20 \times [3\times 3 \times 10 + 1] = 1820)&~\text{[Conv.L2]}\\
    +~& (40 \times [1\times 1 \times 20 + 1] = 840)&~\text{[Conv.L3]}\\
    +~& (32 \times [640 + 1] = 20512)&~\text{[FC.L1]}\\
    +~& (5 \times [32 + 1] = 165)&~\text{[FC.L2]}\\
    =~& 24097
\end{align*}

É possível perceber que, apesar de ser uma arquitetura relativamente pequena, o número de parâmetros as serem aprendidos pode ser grande e cresce consideravelmente quanto maior a arquitetura utilizada.

\subsection{Implementação de CNN para o caso MNIST}

Voltando ao exemplo da classificação de dígitos usando a base MNIST, poderíamos projetar camadas convolucionais de forma a extrair representações com base na estrutura espacial das imagens. Para isso, mostramos na Listagem~\ref{code:mnist3} linhas de código para criar uma camada convolucional: na linha 2 criamos a matriz $\mathbf{W}_1$ utilizando 4 filtros de tamanho $5\times 5 \times 3$, definimos stride 1 na linha 5 e a convolução 2d na linha 6, com uso da função ReLU como ativação na linha 7, produzindo o tensor $\mathbf{x}_2$. 

\noindent \begin{minipage}{\linewidth}
\begin{lstlisting}[caption=Implementação de camadas convolucionais, label=code:mnist3, frame=single]
# inicializacao dos parametros W e B da camada convolucional 1
W1 = tf.Variable(tf.truncated_normal([5, 5, 3, 4], stddev=0.1))
B1 = tf.Variable(tf.ones([4])/10) # 2 e' o numero de mapas de saida da camada

stride = 1  # mantem a imagem no tamanho original
Xconv1 = tf.nn.conv2d(X1, W1, strides=[1, stride, stride, 1], padding='SAME')
X2 = tf.nn.relu(Xconv1 + B1)
\end{lstlisting}
\end{minipage}

Utilizando uma CNN com 3 camadas convolucionais com a seguinte configuração, na sequência, conv1: 8 filtros $5\times 5$ stride 1, conv2: 16 filtros $3\times 3$ stride 2, conv3: 32 filtros $3\times 3$ stride 2, FC1 oculta com 256 neurônios, FC2 de saída com 10 neurônios, é possível alcançar acurácias próximas a $99\%$. No entanto, sempre que ajustamos demais o modelo nos dados de treinamento --- e isso ocorre particularmente em Deep Learning pois o número de parâmetros é muito grande --- há um sério risco do processo de treinamento resultar na memorização dos exemplos que foram utilizados nessa etapa, gerando um efeito conhecido por \textit{overfitting}. Iremos discutir esse ponto na Seção~\ref{sec:conclusions} como umas das limitações dos métodos baseados em DL. Por enquanto veremos como realizar o treinamento da melhor forma possível, buscando evitar esse efeito e atingir uma convergência aceitável do modelo de forma a ser útil nas tarefas que dispomos.

\subsection{Algoritmos de otimização}\label{sec:optimization}

Com uma função de custo definida, precisa-se então ajustar os parâmetros de forma que o custo seja reduzido. Para isso, em geral, usa-se o algoritmo do Gradiente Descendente em combinação com o método de backpropagation, que permite obter o gradiente para a sequência de parâmetros presentes na rede usando a regra da cadeia. Existem diversos materiais disponíveis que apresentam explicações detalhadas sobre o Gradiente Descendente e como funciona o backpropagation. Assim, assume-se que o leitor esteja familiarizado com conceitos fundamentais de ambos os algoritmos, focando em seu funcionamento para CNNs.

% After defining the loss function, we want to adjust the parameters so that the loss is minimized. The Gradient Descent is the standard algorithm for this task, and the backpropagation method is used to obtain the gradient for the sequence of weights using the chain rule. There are many good material and tutorials on both Gradient Descent and the workings of backpropagation. We assume the reader is familiar with the fundamentals of both algorithms, and focus on the particular case of CNNs. 

Como o cálculo de $\mathcal{L}(W)$ é feito baseado em uma amostra finita da base de dados, calcula-se estimativas de Montecarlo da distribuição real que gera os parâmetros. Além disso, é importante lembrar que CNNs possuem muitos parâmetros que precisam ser aprendidos fazendo com que seja necessário treiná-la usando milhares, ou até milhões, de imagens (diversas bases de dados atualmente possuem mais de 1TB de dados). Entretanto, realizar a otimização usando milhões de instâncias torna a utilização de Gradiente Descendente inviável, uma vez que esse algoritmo calcula o gradiente para todas as instâncias individualmente. Essa dificuldade fica clara quando se pensa que para executar uma época (i.e., executar o algoritmo para todas as instâncias na base de treinamento) seria necessário carregar todas as instâncias para uma memória limitada, o que não seria possível. Algumas alternativas foram propostas para superar esse problema. Algumas dessas alternativas são descritas abaixo: SGD, Momentum, AdaGrad, RMSProp e Adam.

% Note that $\mathcal{L}(W)$ is based on a finite dataset and because of that we are computing Montecarlo estimates of the real distribution that generates the parameters. Also, recall that CNNs can have a lot of parameters to be optimized, therefore needing to be trained using thousands or millions of images (many current datasets have more than 1TB of data). But if we have millions of examples to be used in the optimization, then the Gradient Descent is not viable, since this algorithm have to compute the gradient for all examples individually. The difficulty here is easy to see because if we try to run an epoch (i.e. a pass through all the data) we would have to load all the examples into a limited memory, which is not possible. Alternatives to overcome this problem are described below, including the SGD, Momentum, AdaGrad, RMSProp and Adam.

\paragraph{Gradiente Descendente Estocástico (do inglês Stochastic Gradient Descent, SGD)} uma forma de acelerar o treinamento é utilizando métodos que oferecem aproximações do Grandiente Descendente, por exemplo usando amostras aleatórias dos dados ao invés de analisando todas as instâncias existentes. Por esse motivo, o nome desse método é Gradiente Descendente Estocástico, já que ao invés de analisar todos dados disponíveis, analisa-se apenas uma amostra, e dessa forma, adicionando aleatoriedade ao processo. Também é possível calcular o Gradiente Descende usando apenas uma instância por vez (método mais utilizado para analisar fluxos de dados ou aprendizagem online). Na prática, o mais comum é utilizar os chamados \textit{mini-batches} (amostra aleatória dos dados) com um tamanho fixo $B$. Após executar diversas iterações (sendo que cada iteração irá adaptar os parâmetros usando as instâncias no mini-batch atual), espera-se obter uma aproximação do método do Gradiente Descendente.

\begin{equation*}
 W_{t+1} = W_{t} - \eta \sum_{j=1}^{B} \nabla \mathcal{L}(W;x^{\text{B}}_j),
\end{equation*}
na qual $\eta$ é o parâmetro que define a taxa de aprendizado (em inglês \textit{learning rate}), ou seja, esse parâmetro é utilizado para definir o tamanho do passo que será dado na direção apontada pelo gradiente. %Caso $\eta$ seja grande, o algoritmo dará passos maiores na direção do gradiente, enquanto usar uma valor pequeno para $\eta$ resultaria em passos menores na direção do gradiente. 
É comum utilizar valores altos para $\eta$ no começo do treinamento e fazer com ele decaia exponencialmente em função do número de iterações executadas.

% \paragraph{Stochastic Gradient Descent (SGD)} one possible solution to accelerate the process is to use approximate methods that goes through the data in samples composed of random examples drawn from the original dataset. This is why the method is called Stochastic Gradient Descent: now we are not inspecting all available data at a time, but a sample, which adds uncertainty in the process. We can even compute the Gradient Descent using a single example at a time (method often used in streams or online learning). However, in practice it is common to use mini-batches with size $B$. By performing enough iterations (each iteration will compute the new parameters using the examples in the current mini-batch), we assume it is possible to approximate the Gradient Descent method. 
% \begin{equation*}
%  W_{t+1} = W_{t} - \eta \sum_{j=1}^{B} \nabla \mathcal{L}(W;x^{\text{B}}_j),
% \end{equation*}
% in which $\eta$ is the learning rate parameter: a large $\eta$ will produce larger steps in the direction of the gradient, while a small value produces a smaller step in the direction of the gradient. It is common to set a larger initial value for $\eta$, and then exponentially decrease it as a function of the iterations.

Devido ao seu fator aleatório, SGD fornece uma aproximação grosseira do Gradiente Descendente, assim, fazendo com que a convergência não seja suave. Por esse motivo, outras variantes foram propostas buscando compensar esse fato, tais como AdaGrad (\textit{Adaptive Gradient})~\cite{Duchi2011}, AdaDelta (\textit{Adaptive learning rate})~\cite{Zeiler2012adadelta} and Adam (\textit{Adaptive moment estimation})~\cite{Kingma2015}. Tais variantes baseiam-se nos conceitos de momentum e normalização, que serão descritos abaixo.

% In fact SGD is a rough approximation, producing a non-smooth convergence. Because of that, variants were proposed to compensate for that, such as the Adaptive Gradient (AdaGrad)~\cite{Duchi2011}, Adaptive learning rate (AdaDelta)~\cite{Zeiler2012adadelta} and Adaptive moment estimation (Adam)~\cite{Kingma2015}. Those variants basically use the ideas of momentum and normalization, as we describe below. 

\paragraph{Momentum} adiciona um novo hiper-parâmetro que permite controlar a velocidade das mudanças nos parâmetros $W$ da rede. Isso é feito criando um fator de \textit{momentum}, que dá peso para a atual direção do gradiente, e previne que uma nova atualização dos parâmetros $W_{t+1}$ se desvie muito da atual direção de busca no espaço dos parâmetros: 
\begin{align*}
 W_{t+1} = W_{t} &+ \alpha(W_{t} - W_{t-1}) + (1-\alpha) \left[ -\eta \nabla \mathcal{L}(W_t)\right],
\end{align*}
onde $\mathcal{L}(W_t)$ é a perda calculada usando algumas instâncias (usualmente um mini-batch) e os parâmetros atuais $W_t$. É importante notar como o tamanho do passo na iteração $t+1$ agora é limitado pelo passo dado na iteração anterior, $t$.

% \paragraph{Momentum} adds a new variable $\alpha$ to control the change in the parameters $W$. It creates a momentum that prevents the new parameters $W_{t+1}$ from deviating too much from the previous direction:
% \begin{align*}
%  W_{t+1} = W_{t} &+ \alpha(W_{t} - W_{t-1}) + (1-\alpha) \left[ -\eta \nabla \mathcal{L}(W_t)\right],
% \end{align*}
% where $\mathcal{L}(W_t)$ is the loss computed using some examples using the current parameters $W_t$ (often a mini-batch). Note that the magnitude of the step for the iteration $t+1$ now is also constrained by the step taken in the iteration $t$.

\paragraph{AdaGrad} busca dar mais importância a parâmetros pouco utilizados. Isso é feito mantendo um histórico de quanto cada parâmetro influenciou o custo, acumulando os gradientes de forma individual $g_{t+1} = g_{t} + \nabla \mathcal{L}(W_t)^2$. Essa informação é então utilizada para normalizar o passo dado em cada parâmetro:
\begin{equation*}
 W_{t+1} = W_{t} -  \frac{\eta \nabla \mathcal{L}(W_t)^2}{\sqrt{g_{t+1}}+\epsilon},
\end{equation*}
como o gradiente é calculado com base no histórico e para cada parâmetro de forma individual, parâmetros pouco utilizados terão maior influência no próximo passo a ser dado.

% \paragraph{AdaGrad} works by putting more weight on rare or infrequent parameters. It creates a history of how much a given parameter already changed the loss, accumulating the individual gradients $g_{t+1} = g_{t} + \nabla \mathcal{L}(W_t)^2$. Then, the next step is now scaled/normalized for each parameter:
% \begin{equation*}
%  W_{t+1} = W_{t} -  \frac{\eta \nabla \mathcal{L}(W_t)^2}{\sqrt{g_{t+1}}+\epsilon},
% \end{equation*}
% since this historical gradient is computed feature-wise, the infrequent features will have more influence in the next gradient descent step.

\paragraph{RMSProp} calcula médias da magnitude dos gradientes mais recentes para cada parâmetro e as usa para modificar a taxa de aprendizado individualmente antes de aplicar os gradientes. Esse método é similar ao AdaGrad, porém, nele $g_t$ é calculado usando uma média com decaimento exponencial e não com a simples soma dos gradientes:
\begin{align*}
    g_{t+1} &= \gamma g_t + (1-\gamma) \nabla \mathcal{L}(W_t)^2
\end{align*}
$g$ é chamado o momento de segunda ordem de $\nabla \mathcal{L}$ (o termo momento tem a ver com o gradiente não confundir com o efeito \textit{momentum}). 
A atualização dos parâmetros é então feita adicionando-se um tipo de momentum:
\begin{align*}
 W_{t+1} = W_{t} &+ \alpha(W_{t} - W_{t-1}) + (1-\alpha) \left[ - \frac{\eta \nabla \mathcal{L}(W_t)}{\sqrt{g_{t+1}}+\epsilon}\right],
\end{align*}

% \paragraph{RMSProp} computes running averages of recent gradient magnitudes and normalizes using these average so that loosely gradient values are normalized. It is similar to AdaGrad, but here $g_t$ is computed by an exponentially decaying average and not the simple sum of gradients:
% \begin{align*}
%     g_{t+1} &= \gamma g_t + (1-\gamma) \nabla \mathcal{L}(W_t)^2
% \end{align*}
% $g$ is called the second order moment of $\nabla \mathcal{L}$ (don't confuse it with momentum). 
% The final parameter update is given by adding the momentum:
% \begin{align*}
%  W_{t+1} = W_{t} &+ \alpha(W_{t} - W_{t-1}) + (1-\alpha) \left[ - \frac{\eta \nabla \mathcal{L}(W_t)}{\sqrt{g_{t+1}}+\epsilon}\right],
% \end{align*}

\paragraph{Adam} utiliza uma idéia similar ao AdaGrad e ao RMSProp, contudo o \textit{momentum} é usado para tanto para o momento (novamente, não confundir com \textit{momentum}) de primeira quanto o de segunda ordens, tendo assim $\alpha$ e $\gamma$ para controlar $W$ e $g$, respectivamente. A influência de ambos diminui com o tempo de forma que o tamanho do passo diminua conforme aproxima-se do mínimo. Uma variável auxiliar $m$ é usada para facilitar a compreensão:
\begin{align*}
    m_{t+1} &= \alpha_{t+1} g_t + (1-\alpha_{t+1}) \nabla \mathcal{L}(W_t)\\
    \hat m_{t+1} &= \frac{m_{t+1}}{1-\alpha_{t+1}}
\end{align*}
$m$ é o momento de primeira ordem de $\nabla \mathcal{L}$ e $\hat m$ é $m$ após a aplicação do fator de decaimento. Então, precisa-se calcular os gradientes $g$ usado para normalização:
\begin{align*}
    g_{t+1} &= \gamma_{t+1} g_t + (1-\gamma_{t+1}) \nabla \mathcal{L}(W_t)^2\\
    \hat g_{t+1} &= \frac{g_{t+1}}{1-\gamma_{t+1}}
\end{align*}
$g$ é o momento de segunda ordem de $\nabla \mathcal{L}$. A atualização dos parâmetros é então calcula da seguinte forma:
\begin{align*}
    W_{t+1} &= W_t - \frac{\eta \hat m_{t+1}}{\sqrt{\hat g_{t+1}}+\epsilon}
\end{align*}

% \paragraph{Adam} uses an idea that is similar to AdaGrad and RMSProp, but the momentum is used for the first and second order moment so now we have $\alpha$ and $\gamma$ to control the momentum of respectively $W$ and $g$. The influence of both decays over time so that the step size decreases when it approaches minimum. We use a auxiliary variable $m$ for clarity:
% \begin{align*}
%     m_{t+1} &= \alpha_{t+1} g_t + (1-\alpha_{t+1}) \nabla \mathcal{L}(W_t)\\
%     \hat m_{t+1} &= \frac{m_{t+1}}{1-\alpha_{t+1}}
% \end{align*}
% $m$ is called the first order moment of $\nabla \mathcal{L}$ (don't confuse it with momentum) and $\hat m$ is $m$ after applying the decaying factor. Then we need to compute the gradients $g$ to use in the normalization:
% \begin{align*}
%     g_{t+1} &= \gamma_{t+1} g_t + (1-\gamma_{t+1}) \nabla \mathcal{L}(W_t)^2\\
%     \hat g_{t+1} &= \frac{g_{t+1}}{1-\gamma_{t+1}}
% \end{align*}
% $g$ is called the second order moment of $\nabla \mathcal{L}$ (again, don't confuse it with momentum). The final parameter update is given by:
% \begin{align*}
%     W_{t+1} &= W_t - \frac{\eta \hat m_{t+1}}{\sqrt{\hat g_{t+1}}+\epsilon}
% \end{align*}

\subsection{Aspectos importantes no treinamento de CNNs}\label{sec:cnn.training}

\paragraph{Inicialização} a inicialização dos parâmetros é importante para permitir a convergência da rede. Atualmente, se utiliza números aleatórios sorteados a partir de uma distribuição Gaussiana $\mathcal{N}(\mu,\sigma)$ para inicializar os pesos. Pode-se utilizar um valor fixo como o $\sigma=0.01$ utilizado nos nossos exemplos anteriores. Porém o uso desse valor fixo pode atrapalhar a convergência~\cite{Krizhevsky12}. Como alternativa, recomenda-se usar $\mu=0$, $\sigma=\sqrt{2 / n_l}$, onde $n_l$ é o número de conexões na camada $l$, e inicializar vetores bias com valores constantes, conforme fizemos nos nossos exemplos ou ainda iguais a zero~\cite{He2015delving}.

\paragraph{Tamanho do Minibatch} devido ao uso de SGD e suas variantes, é preciso definir o tamanho do minibatch de exemplos a ser utilizado em cada iteração no treinamento. Essa escolha deve levar em consideração restrições de memória mas também os algoritmos de otimização empregados. Um tamanho de batch pequeno pode tornar mais difícil a minimização do custo, mas um tamanho muito grande também pode degradar a velocidade de convergência do SGD para funções objetivo convexas~\cite{Li2014efficient}. 

Apesar disso, foi demonstrado que até certo ponto um valor maior para o tamanho do batch $B$ ajuda a reduzir a variância das atualizações em cada iteração do SGD (pois essa usa a média do custo dos exemplos do batch), permitindo que possamos utilizar tamanhos de passo maiores~\cite{Bottou2016optimization}. Ainda, minibatches maiores são interessantes para o caso de treinamento usando GPUs, gerando maior throughput pelo uso do algoritmo de backpropagation com reuso dos dados por meio da multiplicação entre matrizes (em contrapartida ao uso de várias multiplicações vetor-matriz). Via de regra, escolhe-se $B$ de forma a ocupar toda a memória disponível na GPU, e a maior taxa de aprendizado possível.

Olhando as arquiteturas mais populares (VGGNet~\cite{Simonyan14}, ResNet~\cite{He15}, Inception~\cite{Szegedy2016rethinking,Szegedy2017inception}), notamos de 32 por padrão, mas com até 256 exemplos por batch. Num caso mais extremo, um artigo recente mostro o uso de $B=8192$, o que foi alcançado com uso de 256 GPUs em paralelo e uma  regra de ajuste de escala para a taxa de aprendizado. Com isso os autores conseguiram treinar uma ResNet na base ImageNet em apenas 1 hora~\cite{Goyal2017}.

\paragraph{Regularização} quando usamos uma função de custo para realizar o treinamento de uma rede neural, como apresentado anteriormente, alguns problemas podem surgir. Um caso típico é existência de diversos parâmetros $W$ que façam com que o modelo classifique corretamente o conjunto de treinamento. Como há múltiplas soluções, isso torna mais difícil encontrar bons parâmetros. 

Um termo de regularização adicionado à função de custo auxilia a penalizar essa situação indesejada. A regularização mais comumente utilizada é a de norma L2, que é a soma dos quadrados dos pesos. Como queremos minimizar também essa norma em adição à função de custo, isso desencoraja a existência de alguns poucos valores altos nos parâmetros, fazendo com que as funções aprendidas aproveitem todo o espaço de parâmetros na busca pela solução. Escrevemos a regularização na forma:
\begin{equation*}
\mathcal{L}(W) = \frac{1}{N}\sum_{j=1}^{N} \ell\left( y_j, f(x_j; W)\right) + \lambda R(W).
\end{equation*}

\begin{equation*}
R(W) = \sum_{k}\sum_{l} W_{k,l}^2
\end{equation*}

\noindent onde $\lambda$ é um hiper-parâmetro utilizado para ponderar a regularização, ditando o quanto permitimos que os parâmetros em $W$ possam crescer. Valores para $\lambda$ podem ser encontrados realizando testes por validação cruzada no conjunto de treinamento.

\paragraph{Dropout} é uma técnica para minimizar overfitting proposta em~\cite{Hinton2012improving} que, na fase de treinamento e durante o passo de propagação dos dados pela rede, aleatoriamente desativa com probabilidade $p$ a ativação de neurônios (em particular neurônios de camadas FC). 

Esse procedimento pode ser considerado uma forma de regularização pois, ao desativar aleatoriamente neurônios, perturba os feature maps gerados a cada iteração por meio da redução da complexidade das funções (já que a composição utiliza menos ativações). Ao longo das iterações isso dá um efeito que minimiza a variância e aumenta o viés das funções, e por isso foi demonstrado que o \textit{dropout} tem relações com o método de ensemble Bagging~\cite{Warde2014}. A cada iteração do SGD, cria-se uma rede diferente por meio da subamostragem das ativações. Na fase de teste, no entanto, não se utiliza dropout, e as ativações são re-escaladas com fator $p$ para compensar as ativações que foram desligadas durante a fase de treinamento.

\paragraph{Batch normalization (BN)} as CNNs do estado da arte (em particular Inception~\cite{Szegedy2016rethinking,Szegedy2017inception} e  ResNet~\cite{He15}) utilizam a normalização de batches (BN, Batch normalization)~\cite{Ioffe2015batch}.

Como alternativa, na saída de cada camada  pode-se normalizar o vetor de formas diferentes. O método canal-por-canal normaliza os mapas, considerando cada feature map individualmente ou considerando também mapas vizinhos. Podem ser empregadas normalizações L1, L2 ou variações. Porém esses métodos foram abandonados em favor do BN.

BN também tem efeito regularizador, normalizando as ativações da camada anterior em cada batch de entrada, mantendo a ativação média próxima a 0 (centralizada) e o desvio padrão das ativações próximo a 1, e utilizando parâmetros $\gamma$ e $\beta$ para compor a transformação linear do vetor normalizado:
\begin{align}
    \operatorname{BN}_{\gamma,\beta}(x_i) = \gamma\left(\frac{x_i-\mu_B}{\sqrt{\sigma^2_B+\epsilon}} \right)+\beta.
\end{align}
Note que $\gamma$ e $\beta$ podem ser incluídos no modelo de forma a serem ajustados/aprendidos durante o backpropagation~\cite{Ioffe2015batch}, o que pode ajustas a normalização a partir dos dados, e até mesmo cancelar essa normalização, i.e. se esses forem ajustados para $\gamma=\sqrt{\sigma^2_B}$ e $\beta=\mu_B$.

BN se tornou um método padrão nos últimos anos, tendo substituído o uso de regularização e em alguns casos até mesmo tornando desnecessário o uso de dropout.

\paragraph{Data-augmentation} como mencionado anteriormente, redes profundas possuem um espaço de parâmetros muito grande a ser otimizado. Isso faz com que seja necessário dispor de um número muitas vezes proibitivo de exemplos rotulados. Assim, podem ser empregados métodos de \textit{data augmentation}, i.e. geração de uma base de treinamento aumentada. Isso é feito comumente aplicando operações de processamento de imagens em cada imagem da base gerando 5, 10 (o um mais) novas imagens por imagem original existente.

Alguns métodos para geração de imagens incluem~\cite{Chatfield14}: (i) corte da imagem de entrada em diferentes posições --- tirando vantagem do fato de que as imagens possuem resolução superior à entrada comumente esperada pelas CNNs (e.g. $224\times 224$), ao invés de redimensionarmos uma imagem para esse valor, cortamos a imagem original de maior resolução em diversas subimagens de tamanho $224\times 224$; (ii) girar a imagem horizontalmente, e também verticalmente caso faça sentido no contexto das imagens de entrada, como por exemplo é o caso de imagens astronômicas, de sensoriamento remoto, etc.; (iii) adicionando ruído às imagens de entrada~\cite{Nazare2017}; (iv) criando imagens por meio da aplicação de Análise de Componentes Principais (PCA) nos canais de cor, como no método Fancy PCA~\cite{Krizhevsky12}.

\paragraph{Pre-processamento} é possível pré-processar as imagens de entrada de diversas formas. As mais comuns incluem: (i) computar a imagem média para todo o conjunto de treinamento e subtrair essa média de cada imagem; (ii) normalização $z$-score, (iii) PCA \textit{whitening} que primeiramente tenta descorrelacionar os dados projetando os dados originais centrados na origem em uma auto-base (em termos dos auto-vetores), e então dividindo os dados nessa nova base pelos auto-valores relacionados de maneira a normalizar a escala.

O método $z$-score é o mais utilizado dentre os citados (centralização por meio da subtração da média, e normalização por meio da divisão pelo desvio padrão), o que pode ser visto como uma forma de realizar \textit{whitening} ~\cite{Lecun2012efficient}. 

\subsection{Utilizando modelos pré-treinados: fine-tuning e extração de características} 

Comumente dispomos de um conjunto de dados pequeno, inviáveis para permitir o treinamento de uma CNN a partir do zero, mesmo empregando métodos de data augmentation (note que a geração de imagens apenas cria versões perturbadas das mesmas imagens). Nesse caso é muito útil utilizar um modelo cujos parâmetros já foram encontrados para um conjunto de dados grande (como por exemplo para a base ImageNet~\cite{Deng09} que possui mais de um milhão de imagens de 1000 classes). 

O processo de \textbf{fine-tuning} consiste basicamente de continuar o treinamento a partir dos pesos iniciais, mas agora utilizando um subconjunto de sua base de dados. Note que é provável que essa nova base de dados tenha outras classes (em contrapartida por exemplo às 1000 classes da ImageNet). Assim se você deseja criar um novo classificador será preciso remover a última camada e adicionar uma nova camada de saída com o número de classes desejado, a qual deverá ser treinada do zero a partir de uma inicialização aleatória.

A partir disso, há várias abordagens para realizar fine-tuning, que incluem por exemplo (i) permitir que o algoritmo ajuste todos os pesos da rede com base nas novas imagens, (ii) congelar algumas camadas e permitir que o algoritmo ajuste apenas os parâmetros um subconjunto de camadas -- por exemplo podemos ajustar apenas os pesos da última camada criada, ou apenas os pesos das FCs, etc, (iii) criar novas camadas adicionais com números e tamanhos de filtros diferentes.

O mais comum é a abordagem (ii), congelando das primeiras camadas (em geral convolucionais) e permitindo que as camadas mais profundas se adaptem.

Para obter \textbf{extração de características}, mesmo sem realizar fine-tuning, oferece-se como entrada as imagens desejadas, e utilizamos como vetor de características a saída de uma das camadas da rede (antes da camada de saída). Comumente se utiliza a penúltima camada: por exemplo na VGGNet~\cite{Simonyan14} a FC2 tem 4096 neurônios, gerando um vetor de 4096 elementos, já na Inception V3~\cite{Szegedy2016rethinking} temos 2048 elementos na penúltima camada. Caso a dimensionalidade seja alta, é possível utilizar algum método de redução de dimensionalidade ou quantização baseada por exemplo em PCA~\cite{Ponti2016image} ou Product Quantization~\cite{Bui2017compact}.

\section{Porque Deep Learning funciona?}
\label{sec:why}

Ao longo desse capítulo esperamos que o leitor tenha compreendido com clareza o funcionamento dos métodos de DL, incluindo os componentes básicos das redes profundas (tipos de camadas, representações, funções de ativação, funções de custo etc.) e os algoritmos e técnicas fundamentais (otimização, regularização, normalização, etc.). Mostramos que, combinando uma arquitetura adequada em termos do número de camadas, quantidade de neurônios por camada, e atentando para detalhes no processo de otimização, é possível alcançar excelentes resultados na regressão e classificação de sinais de alta dimensionalidade. No entanto resta a dúvida: porque esses métodos funcionam tão bem? É justificada a grande atenção atualmente dispensada a esses métodos, e a expectativa que os cerca em termos da solução de outros problemas? Quais as garantias teóricas que embasam seu funcionamento?

Nas seções anteriores mostramos que a hipótese para o sucesso dos métodos de DL está na composição de funções, que realiza transformações sucessivas a partir do vetor de entrada. No entanto, analisar esses algoritmos é difícil devido à falta de estrutura para compreensão das suas invariantes, bem como das transformações intrínsecas.

\paragraph{O papel da profundidade}: há alguns estudos que se dedicam a demonstrar porque redes profundas são melhores do que redes superficiais a partir de teoremas de aproximação. Um dos trabalhos nessa direção é o de Kolmogorov (1936)~\cite{Kolmogorov1956representation} que demonstrou um limite superior para a aproximação de funções contínuas por meio de subespaços de funções, seguido por Warren (1968) que encontrou limites para a aproximação por meio de polinômios, em termos de graus polinomiais e dimensão da função alvo, provando a dimensão VC de polinômios~\cite{Warren1968lower}. O refinamento desses resultados veio com teoremas de hierarquia de profundidade em complexidade de circuitos~\cite{Hastad1986}, demonstrando a diferença entre circuitos com uma certa profundidade de circuitos de menor profundidade. Esse resultado é uma das bases para o estudo de redes neurais profundas~\cite{Mhaskar2016deep}~\cite{Telgarsky2016benefits}~\cite{Lin2016does}.

Entre os resultados obtidos~\cite{Lin2016does} demonstra que o custo de achatarmos o grafo formado pela rede neural para um determinado problema é comumente proibitivo, relacionando o uso de múltiplas camadas com algoritmos de divisão e conquista. Um exemplo apontado pelos autores é a Transformada Rápida de Fourier (FFT), que basicamente faz uso de uma fatorização esparsa da matriz da Transformada Discreta de Fourier (DFT), resultando em $O(n \log n)$ elementos não-nulos na matrix, em contraste com os $n^2$ elementos não nulos na matriz da DFT. Assim, se imaginarmos a FFT como operações organizadas em um grafo (rede), achatar as operações em uma única camada irá aumentar a contagem de $n\log n$ para $n^2$.

Telgarsky (2016)~\cite{Telgarsky2016benefits} vai além das analogias e mostra que para qualquer inteiro positivo $k$, existem redes neurais com $\Theta(k^3)$ camadas, cada qual com $\Theta(1)$ nós e $\Theta(1)$ parâmetros distintos, que não podem ser aproximadas por redes neurais com $O(k)$ camadas, a não ser que essas tenham uma quantidade de neurônios exponencialmente grande, da ordem de $\Omega(2^k)$ neurônios. Esse resultado é provado para portas semi-algébricas, as quais incluem funções como ReLU, operadores de máximo (como \textit{maxpooling}), e funções polinomiais por partes. Isso permitiu estabelecer o papel fundamental da profundidade tanto em redes neurais convencionais quanto convolucionais que utilizam ReLU e operações de máximo. Abaixo a reprodução do teorema.

\begin{thm} Seja qualquer inteiro $k\geq 1$ e qualquer dimensão $d\geq 1$. Existe uma função $f: \mathbb{R}^d \rightarrow \mathbb{R}$ computada por uma rede neural com portas ReLU em $2k^3+8$ camadas, com $3k^3+12$ neurônios no total, e $4+d$ parâmetros distintos de forma que:
\begin{equation*}
\inf_{g\in \mathcal{C}} \int_{[0,1]^d} |f(x) - g(x)| dx \geq \frac{1}{64},
\end{equation*}
onde $\mathcal{C}$ é a união dos seguintes dois conjuntos de funções:\\
(1) funções computadas por redes compostas por portas $(t,\alpha,\beta)$-semi-algébricas com número de camadas $\leq k$ e número de neurônios $2^k/(t\alpha \beta)$ --- como é o caso de redes que utilizam funções ReLU ou redes convolucionais com funções ReLU e portas de maximização;\\
(2) funções computadas por combinações lineares de um número de árvores de decisão $\leq t$ com $2^{k^3} / t$ neurônios --- como as funções utilizadas em boosted decision trees.
\end{thm}

O nome porta semi-algébrica vem de conjuntos semi-algébricos, definidos por uniões e intersecções de desigualdades polinomiais, e denota uma função que mapeia algum domínio em $\mathbb{R}$. Em redes neurais o domínio da função deve ser particionado em três argumentos: o vetor de entrada entrada, o vetor de parâmetros, e o vetor de números reais advindos de neurônios predecessores.

A prova do Teorema é feita em três etapas, demonstrando: (1) que funções com poucas oscilações aproximam de forma pobre funções com mais oscilações, (2) que funções computadas por redes neurais com menos camadas possuem menos oscilações, (3) que funções computadas por redes com mais camadas podem ter mais oscilações. A prova vem do teorema de hierarquia de profundidade para circuitos booleanos do trabalho seminal de H\r{a}stad~\cite{Hastad1986} que estabelece não ser possível aproximar funções de paridade de circuitos profundos por circuitos menos profundos a não ser que o tamanho desses últimos seja exponencial. Por questões de espaço recomendamos a leitura do artigo completo de Telgarsky~\cite{Telgarsky2016benefits}.

\paragraph{Propriedades de contração e separação}
Segundo Mallat (2016)~\cite{Mallat2016understanding}, o funcionamento das CNN está ligado à eliminação de elementos/variáveis não informativas, via contração ou redução da dimensão espacial nas direções mais apropriadas a um certo problema. O autor define o problema de aprendizado supervisionado, que computa uma aproximação $\hat f(\mathbf{x})$ de uma função  $f(\mathbf{x})$ por meio de um conjunto de treinamento $\mathbf{X} \in \Omega$, sendo que o domínio de $\Omega$ é um subconjunto aberto de alta dimensionalidade de $\mathbb{R}^d$ (e não um manifold de baixa dimensão). Então os conceitos de separação e linearização são apresentados.

Idealmente, deseja-se reduzir a dimensão de um vetor de entrada $\mathbf{x}$ computando um vetor de menor dimensão $\phi(\mathbf{x})$. Assim,  $\phi()$ é um operador de contração que reduz o intervalo de variações de $\mathbf{x}$ e que, ao mesmo tempo, separa diferentes valores da função $f$ (classificador ou regressor), ou seja, $\phi(\mathbf{x}) \leq \phi(\mathbf{x'})$ se $f(\mathbf{x}) \leq f(\mathbf{x'})$. Assim, $\phi$ \textbf{separa} $f$, encontrando uma projeção linear de $\mathbf{x}$ em algum espaço $\mathcal{V}$ de menor dimensão $k$, que separa $f$.

Como estratégia alternativa, as variações de $f$ podem ser linearizadas de forma a reduzir a dimensionalidade. Nesse cenário um projetor linear de baixa dimensionalidade de $\mathbf{x}$ pode separar os valores de $f$ se essa permanecer constante na direção do espaço linear de alta dimensão. Após encontrar $\phi(\mathbf{x})$ que realize essa linearização mantendo $f(\mathbf{x})$ constante, a dimensão é reduzida aplicando projeção linear em $\phi(\mathbf{x})$. 

Redes neurais profundas progressivamente contraem o espaço e lineariza transformações ao longo das quais $f$ permanece aproximadamente constante, de forma a preservar a separação. A combinação de canais (feature maps) provêm a flexibilidade necessária para estender translações para maiores grupos de simetrias locais. Assim, as redes são estruturadas de forma a fatorizar grupos de simetrias, em que todos os operadores lineares são convoluções.

Assim como em~\cite{Liu2016learning}, Mallat~\cite{Mallat2016understanding} faz conexões com conceitos de física, demonstrando que redes hierárquicas multiescala computam convoluções de forma que as classes sejam preservadas. Os pesos geram repostas fortes a padrões particulares e são invariantes a uma série de transformações e os filtros são adaptados para produzir representações esparsas de vetores de suporte multiescala. Esses vetores proveriam então um código distribuído que define uma memorização de padrões invariante. A interpretação de geometria diferencial é a de que os pesos são transportados ao longo de um maço de fibras (conceito matemático que estaria relacionado a atualização dos filtros). Conforme a contração aumenta, aumenta o número de vetores de suporte necessários para realizar a separação o que indicaria um incremento também no número de fibras permitindo explicar simetrias e invariâncias suficientemente estáveis para a obtenção de transferência de aprendizado. Isso possibilita importante interpretação do aprendizado e abre espaço para pesquisas teóricas na área.

\subsection{Limitações de Deep Learning e Considerações Finais}
\label{sec:conclusions}

Técnicas de Deep Learning tem conseguido atingir resultados estatisticamente impressionantes em particular, como demonstrado, pelo uso de múltiplas camadas. Entretanto, existem limitações no uso de redes neurais profundas pois essas são basicamente uma forma de aprender uma série de transformações a serem aplicadas ao vetor de entrada.  Essas transformações são dadas por um grande conjunto de pesos (parâmetros) que são atualizados durante a etapa de treinamento de forma a minimizar a função de custo. A primeira limitação para que seja possível realizar o treinamento, é que tais transformações precisam ser deriváveis, isso é, o mapa entre a entrada e saída da rede deve ser contínuo e idealmente suave~\cite{Chollet2017}.

\begin{figure}
\begin{center}
 \includegraphics[width=0.7\linewidth]{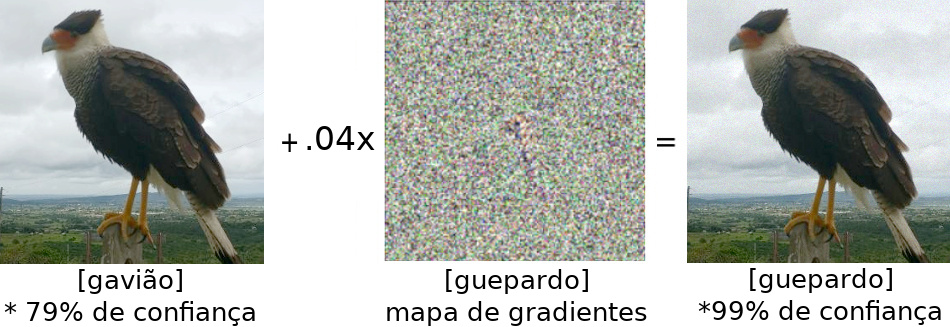}
\end{center}
 \caption{Imagem gerada pela adição de 4\% do gradiente de features de guepardo em uma imagem de um gavião, que é então classificada como guepardo com alta confiança pela CNN.}
 \label{fig:adversarial2}
\end{figure}

A segunda limitação diz respeito à abstração e adaptação: redes neurais profundas precisam de  quantidades massivas de dados rotulados para aprender conceitos simples; em contrapartida, seres humanos são capazes de aprender um conceito a partir de pequenas quantidades de exemplos. No problema de classificação de dígitos, é necessário fornecer milhares de exemplos de dígitos, escritos por pessoas diferentes para que a rede possa aprender a diferenciá-los, enquanto um humano seria capaz de aprender os mesmos conceitos com alguns poucos exemplos escritos por uma única pessoa. Humanos também conseguem adaptar facilmente esses conceitos com o passar do tempo. 

Mesmo considerando tarefas a princípio solucionáveis via Deep Learning, existem diversos estudos mostrando falhas que fazem duvidar da real capacidade de generalização desses modelos, em particular: (i) perturbações nos dados de entrada, tais como ruído, podem impactar significativamente os resultados~\cite{Nazare2017}, (ii) o uso de imagens invertidas na fase de teste faz com que as redes errem completamente os resultados~\cite{Hosseini2017deep}, (iii) é possível criar imagens que visualmente parecem conter somente ruído aleatório, mas que fazem com que CNNs as classifiquem com $99\%$ de certeza~\cite{Nguyen2015deep}, (iv) é possível desenvolver perturbações imperceptíveis a humanos (em torno de 4\%) que fazem redes produzirem saídas completamente erradas~\cite{Papernot2016limitations}, conforme mostramos na Figura~\ref{fig:adversarial2}, um exemplo adaptado adicionando informações do gradiente de de uma imagem de guepardo a uma imagem de um falcão, o que faz com que a imagem seja classificada com alta confiança como guepardo por uma CNN do estado da arte. Para aliviar o impacto dessa fragilidade dos métodos de Deep Learning a ataques, estudos recentes recomendam a inclusão de exemplos adversários no conjunto de treinamento ou o aumento da capacidade (em termos de filtros/neurônios por camada), aumentando o número de parâmetros e assim a complexidade do espaço de funções permitidas pelos modelo~\cite{Rozsa2016accuracy},\cite{Madry2017towards}. 

Ao analizar os cenários citados do ponto de vista da Teoria do Aprendizado Estatístico~\cite{Vapnik2013nature}, podemos criar a hipótese de que esses métodos estão de fato aprendendo um modelo baseado em memória. Eles funcionariam bem em diversos casos pois são treinados em milhões de exemplos, fazendo com que imagens nunca vistas se encaixem nas informações memorizadas. Estudos recentes incluem a análise tensorial de representações~\cite{Cohen2016expressive} e incertezas~\cite{Gal2016dropout} para tentar esclarecer o comportamento do aprendizado das redes neurais profundas. Ainda~\cite{Mallat2016understanding} e~\cite{Lin2016does} publicaram resultados teóricos importantes, com interpretações que podem dar novas direções à compreensão e ao estudo das garantias teóricas e limites, mas ainda há muito a percorrer do ponto de vista do embasamento teórico relacionado ao aprendizado supervisionado e não supervisionado utilizando redes profundas. 

Para maiores detalhes, aplicações, e outros modelos de redes incluindo Autoencoders e Modelos Gerativos sugerimos a leitura do survey recente sobre Deep Learning em Visão Computacional~\cite{Ponti2017everything}, bem como o livro de Goodfelow et al. disponível online~\cite{Goodfellow16}.

É inegável o impacto positivo de Deep Learning para diversas áreas em que o aprendizado de máquina é empregado para busca de soluções. Compreendendo suas limitações e funcionamento, é possível continuar a explorar esses métodos e alcançar avanços notáveis em diversas aplicações. 

\section*{Agradecimentos}
Os autores agradecem o apoio da FAPESP (processos \#2016/16411-4 e \#2015/05310-3)

\bibliographystyle{plain}
\bibliography{deepLearning}

\begin{thebibliography}{10}

\bibitem{BenDavid2014}
Shai Ben-David and Shai Shalev-Shwartz.
\newblock {\em Understanding Machine Learning: from theory to algorithms}.
\newblock Cambridge, 2014.

\bibitem{Bengio2013RepresentationLearningReview}
Yoshua Bengio, Aaron Courville, and Pascal Vincent.
\newblock Representation learning: A review and new perspectives.
\newblock {\em IEEE Trans. Pattern Anal. Mach. Intell.}, 35(8):1798--1828,
  August 2013.

\bibitem{Bottou2016optimization}
L{\'e}on Bottou, Frank~E Curtis, and Jorge Nocedal.
\newblock Optimization methods for large-scale machine learning.
\newblock {\em arXiv preprint arXiv:1606.04838}, 2016.

\bibitem{Bui2017compact}
T~Bui, L~Ribeiro, M~Ponti, and John Collomosse.
\newblock Compact descriptors for sketch-based image retrieval using a triplet
  loss convolutional neural network.
\newblock {\em Computer Vision and Image Understanding}, 2017.

\bibitem{Chatfield14}
Ken Chatfield, Karen Simonyan, Andrea Vedaldi, and Andrew Zisserman.
\newblock Return of the devil in the details: Delving deep into convolutional
  nets.
\newblock In {\em British Machine Vision Conference (BMVC)}, page 1405.3531,
  2014.

\bibitem{Chollet2017}
F.~Chollet.
\newblock {\em Deep Learning with Python}.
\newblock Manning, 2017.

\bibitem{Cohen2016expressive}
Nadav Cohen, Or~Sharir, and Amnon Shashua.
\newblock On the expressive power of deep learning: A tensor analysis.
\newblock In {\em Conference on Learning Theory}, pages 698--728, 2016.

\bibitem{Deng09}
J.~Deng, W.~Dong, R.~Socher, L.-J. Li, K.~Li, and L.~Fei-Fei.
\newblock {ImageNet: A Large-Scale Hierarchical Image Database}.
\newblock In {\em CVPR09}, 2009.

\bibitem{Duchi2011}
John Duchi, Elad Hazan, and Yoram Singer.
\newblock Adaptive subgradient methods for online learning and stochastic
  optimization.
\newblock {\em Journal of Machine Learning Research}, 12(Jul):2121--2159, 2011.

\bibitem{Fischer2014training}
Asja Fischer and Christian Igel.
\newblock Training restricted boltzmann machines: An introduction.
\newblock {\em Pattern Recognition}, 47(1):25--39, 2014.

\bibitem{Fukushima88}
Kunihiko Fukushima.
\newblock Neocognitron: A hierarchical neural network capable of visual pattern
  recognition.
\newblock {\em Neural networks}, 1(2):119--130, 1988.

\bibitem{Gal2016dropout}
Yarin Gal and Zoubin Ghahramani.
\newblock Dropout as a bayesian approximation: Representing model uncertainty
  in deep learning.
\newblock In {\em International Conference on Machine Learning}, pages
  1050--1059, 2016.

\bibitem{Gonzalez2007}
Rafael Gonzalez and Richard Woods.
\newblock {\em Digital Image Processing}.
\newblock Pearson, 3 edition, 2007.

\bibitem{Goodfellow16}
Ian Goodfellow, Yoshua Bengio, and Aaron Courville.
\newblock {\em Deep Learning}.
\newblock MIT Press, 2016.
\newblock \url{http://www.deeplearningbook.org}.

\bibitem{Goodfellow14}
Ian~J. Goodfellow, Jean Pouget-Abadie, Mehdi Mirza, Bing Xu, David
  Warde-Farley, Sherjil Ozair, Aaron Courville, and Yoshua Bengio.
\newblock Generative adversarial nets.
\newblock In {\em Advances in Neural Information Processing Systems 27}, 2014.

\bibitem{Goyal2017}
P.~Goyal, P.~Dollar, R.~Girshick, P.~Noordhuis, L.~Wesolowski, A.~Kyrola,
  A.~Tulloch, Y.~Jia, and K.~He.
\newblock Accurate, large minibatch {SGD}: Training imagenet in 1 hour.

\bibitem{Graves2013speech}
Alex Graves, Abdel-rahman Mohamed, and Geoffrey Hinton.
\newblock Speech recognition with deep recurrent neural networks.
\newblock In {\em {IEEE} International Conference on Acoustics, Speech and
  Signal Processing (ICASSP)}, pages 6645--6649, 2013.

\bibitem{He15}
Kaiming He, Xiangyu Zhang, Shaoqing Ren, and Jian Sun.
\newblock Deep residual learning for image recognition.
\newblock {\em CoRR}, abs/1512.03385, 2015.

\bibitem{He2015delving}
Kaiming He, Xiangyu Zhang, Shaoqing Ren, and Jian Sun.
\newblock Delving deep into rectifiers: Surpassing human-level performance on
  imagenet classification.
\newblock In {\em Proceedings of the IEEE international conference on computer
  vision}, pages 1026--1034, 2015.

\bibitem{Hinton2012improving}
Geoffrey~E Hinton, Nitish Srivastava, Alex Krizhevsky, Ilya Sutskever, and
  Ruslan~R Salakhutdinov.
\newblock Improving neural networks by preventing co-adaptation of feature
  detectors.
\newblock {\em arXiv preprint arXiv:1207.0580}, 2012.

\bibitem{Hosseini2017deep}
Hossein Hosseini and Radha Poovendran.
\newblock Deep neural networks do not recognize negative images.
\newblock {\em arXiv preprint arXiv:1703.06857}, 2017.

\bibitem{Hastad1986}
Johan Håstad.
\newblock {\em Computational Limitations of Small Depth Circuits}.
\newblock PhD thesis, MIT, 1986.

\bibitem{Ioffe2015batch}
Sergey Ioffe and Christian Szegedy.
\newblock Batch normalization: Accelerating deep network training by reducing
  internal covariate shift.
\newblock In {\em International Conference on Machine Learning}, pages
  448--456, 2015.

\bibitem{Kingma2015}
Diederik Kingma and Jimmy Ba.
\newblock Adam: A method for stochastic optimization.
\newblock In {\em 3rd International Conference on Learning Representations
  (ICLR)}, 2015.

\bibitem{Kolmogorov1956representation}
AN~Kolmogorov.
\newblock The representation of continuous functions of several variables by
  superpositions of continuous functions of a smaller number of variables.
\newblock {\em Doklady Akademii Nauk SSSR}, 108(2):179--182, 1956.

\bibitem{Krizhevsky12}
Alex Krizhevsky, Ilya Sutskever, and Geoffrey~E. Hinton.
\newblock Imagenet classification with deep convolutional neural networks.
\newblock In {\em Advances in Neural Information Processing Systems 25: 26th
  Annual Conference on Neural Information Processing Systems.}, pages
  1106--1114, 2012.

\bibitem{Larochelle2012}
Hugo Larochelle, Michael Mandel, Razvan Pascanu, and Yoshua Bengio.
\newblock Learning algorithms for the classification restricted boltzmann
  machine.
\newblock {\em Journal of Machine Learning Research}, 13(Mar):643--669, 2012.

\bibitem{Lecun98}
Yann LeCun, L{\'e}on Bottou, Yoshua Bengio, and Patrick Haffner.
\newblock Gradient-based learning applied to document recognition.
\newblock {\em Proceedings of the IEEE}, 86(11):2278--2324, 1998.

\bibitem{Lecun2012efficient}
Yann~A LeCun, L{\'e}on Bottou, Genevieve~B Orr, and Klaus-Robert M{\"u}ller.
\newblock Efficient backprop.
\newblock In {\em Neural networks: Tricks of the trade}, pages 9--48. Springer,
  2012.

\bibitem{Li2014efficient}
Mu~Li, Tong Zhang, Yuqiang Chen, and Alexander~J Smola.
\newblock Efficient mini-batch training for stochastic optimization.
\newblock In {\em Proceedings of the 20th ACM SIGKDD international conference
  on Knowledge discovery and data mining}, pages 661--670. ACM, 2014.

\bibitem{Lin2016does}
Henry~W Lin, Max Tegmark, and David Rolnick.
\newblock Why does deep and cheap learning work so well?
\newblock {\em Journal of Statistical Physics}, pages 1--25, 2016.

\bibitem{Liu2016learning}
Fayao Liu, Chunhua Shen, Guosheng Lin, and Ian Reid.
\newblock Learning depth from single monocular images using deep convolutional
  neural fields.
\newblock {\em IEEE Transactions on Pattern Analysis and Machine Intelligence},
  38(10):2024--2039, 2016.

\bibitem{Madry2017towards}
Aleksander Madry, Aleksandar Makelov, Ludwig Schmidt, Dimitris Tsipras, and
  Adrian Vladu.
\newblock Towards deep learning models resistant to adversarial attacks.
\newblock {\em arXiv preprint arXiv:1706.06083}, 2017.

\bibitem{Mallat2016understanding}
St{\'e}phane Mallat.
\newblock Understanding deep convolutional networks.
\newblock {\em Phil. Trans. R. Soc. A}, 374(2065):20150203, 2016.

\bibitem{Mhaskar2016deep}
H.N. Mhaskar and T.~Poggio.
\newblock Deep vs. shallow networks: An approximation theory perspective.
\newblock {\em Analysis and Applications}, 14(06):829--848, 2016.

\bibitem{Nair10}
Vinod Nair and Geoffrey~E. Hinton.
\newblock Rectified linear units improve restricted boltzmann machines.
\newblock In Johannes Fürnkranz and Thorsten Joachims, editors, {\em
  Proceedings of the 27th International Conference on Machine Learning
  (ICML-10)}, pages 807--814. Omnipress, 2010.

\bibitem{Nazare2017}
T.~Nazare, G.~Paranhos~da Costa, W.~Contato, and M.~Ponti.
\newblock Deep convolutional neural networks and noisy images.
\newblock In {\em Iberoamerican Conference on Pattern Recognition (CIARP)},
  2017.

\bibitem{Nguyen2015deep}
Anh Nguyen, Jason Yosinski, and Jeff Clune.
\newblock Deep neural networks are easily fooled: High confidence predictions
  for unrecognizable images.
\newblock In {\em IEEE Conference on Computer Vision and Pattern Recognition
  (CVPR)}, pages 427--436, 2015.

\bibitem{Papernot2016limitations}
Nicolas Papernot, Patrick McDaniel, Somesh Jha, Matt Fredrikson, Z~Berkay
  Celik, and Ananthram Swami.
\newblock The limitations of deep learning in adversarial settings.
\newblock In {\em IEEE European Symposium on Security and Privacy (EuroS\&P)},
  pages 372--387. IEEE, 2016.

\bibitem{Pascanu2013difficulty}
Razvan Pascanu, Tomas Mikolov, and Yoshua Bengio.
\newblock On the difficulty of training recurrent neural networks.
\newblock In {\em International Conference on Machine Learning}, pages
  1310--1318, 2013.

\bibitem{Ponti2016imagerestoration}
Moacir Ponti, Elias~S Helou, Paulo Jorge~SG Ferreira, and Nelson~DA
  Mascarenhas.
\newblock Image restoration using gradient iteration and constraints for band
  extrapolation.
\newblock {\em IEEE Journal of Selected Topics in Signal Processing},
  10(1):71--80, 2016.

\bibitem{Ponti2017decision}
Moacir Ponti, Josef Kittler, Mateus Riva, Te{\'o}filo de~Campos, and Cemre Zor.
\newblock A decision cognizant {K}ullback--{L}eibler divergence.
\newblock {\em Pattern Recognition}, 61:470--478, 2017.

\bibitem{Ponti2016image}
Moacir Ponti, Tiago~S Nazar{\'e}, and Gabriela~S Thum{\'e}.
\newblock Image quantization as a dimensionality reduction procedure in color
  and texture feature extraction.
\newblock {\em Neurocomputing}, 173:385--396, 2016.

\bibitem{Ponti2017everything}
Moacir Ponti, Leonardo~S.F. Ribeiro, Tiago~S. Nazare, Tu~Bui, and John
  Collomosse.
\newblock Everything you wanted to know about deep learning for computer vision
  but were afraid to ask.
\newblock In {\em SIBGRAPI --- Conference on Graphics, Patterns and Images
  Tutorials (SIBGRAPI-T 2017)}, pages 1--26, 2017.

\bibitem{Rozsa2016accuracy}
Andras Rozsa, Manuel G{\"u}nther, and Terrance~E Boult.
\newblock Are accuracy and robustness correlated.
\newblock In {\em Machine Learning and Applications (ICMLA), 2016 15th IEEE
  International Conference on}, pages 227--232. IEEE, 2016.

\bibitem{ILSVRC15}
Olga Russakovsky, Jia Deng, Hao Su, Jonathan Krause, Sanjeev Satheesh, Sean Ma,
  Zhiheng Huang, Andrej Karpathy, Aditya Khosla, Michael Bernstein,
  Alexander~C. Berg, and Li~Fei-Fei.
\newblock {ImageNet Large Scale Visual Recognition Challenge}.
\newblock {\em International Journal of Computer Vision (IJCV)},
  115(3):211--252, 2015.

\bibitem{Salakhutdinov2009deep}
Ruslan Salakhutdinov and Geoffrey Hinton.
\newblock Deep boltzmann machines.
\newblock In {\em Artificial Intelligence and Statistics}, pages 448--455,
  2009.

\bibitem{Simonyan14}
K.~Simonyan and A.~Zisserman.
\newblock Very deep convolutional networks for large-scale image recognition.
\newblock {\em CoRR}, abs/1409.1556, 2014.

\bibitem{Springenberg2015striving}
Jost~Tobias Springenberg, Alexey Dosovitskiy, Thomas Brox, and Martin
  Riedmiller.
\newblock Striving for simplicity: The all convolutional net.
\newblock In {\em ICLR (workshop track)}, 2015.

\bibitem{Szegedy2017inception}
Christian Szegedy, Sergey Ioffe, Vincent Vanhoucke, and Alexander~A Alemi.
\newblock Inception-v4, inception-resnet and the impact of residual connections
  on learning.
\newblock In {\em AAAI}, pages 4278--4284, 2017.

\bibitem{Szegedy2016rethinking}
Christian Szegedy, Vincent Vanhoucke, Sergey Ioffe, Jon Shlens, and Zbigniew
  Wojna.
\newblock Rethinking the inception architecture for computer vision.
\newblock In {\em Proceedings of the IEEE Conference on Computer Vision and
  Pattern Recognition}, pages 2818--2826, 2016.

\bibitem{Telgarsky2016benefits}
Matus Telgarsky.
\newblock Benefits of depth in neural networks.
\newblock {\em arXiv preprint arXiv:1602.04485}, 2016.

\bibitem{Vapnik2013nature}
Vladimir Vapnik.
\newblock {\em The nature of statistical learning theory}.
\newblock Springer science \& business media, 2013.

\bibitem{Warde2014}
David Warde-Farley, Ian~J Goodfellow, Aaron Courville, and Yoshua Bengio.
\newblock An empirical analysis of dropout in piecewise linear networks.
\newblock In {\em ICLR 2014}, 2014.

\bibitem{Warren1968lower}
Hugh~E Warren.
\newblock Lower bounds for approximation by nonlinear manifolds.
\newblock {\em Transactions of the American Mathematical Society},
  133(1):167--178, 1968.

\bibitem{Zeiler2012adadelta}
Matthew~D Zeiler.
\newblock Adadelta: an adaptive learning rate method.
\newblock {\em arXiv preprint arXiv:1212.5701}, 2012.

\bibitem{Zheng2015conditional}
Shuai Zheng, Sadeep Jayasumana, Bernardino Romera-Paredes, Vibhav Vineet,
  Zhizhong Su, Dalong Du, Chang Huang, and Philip Torr.
\newblock Conditional random fields as recurrent neural networks.
\newblock In {\em International Conference on Computer Vision (ICCV)}, 2015.

\end{thebibliography}

\section*{Sobre os Autores}

\footnotesize
\begin{wrapfigure}{l}{0.2\linewidth}
\includegraphics[width=1\linewidth]{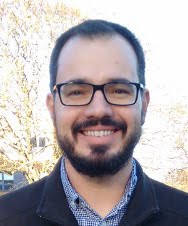}
\end{wrapfigure}
\paragraph{Moacir A. Ponti} é professor Doutor no Instituto de Ciências Matemáticas e de Computação (ICMC) da Universidade de São Paulo (USP), em São Carlos/SP, Brasil. Possui Doutorado (2008) e Mestrado (2004) pela Universidade Federal de São Carlos. Foi pesquisador visitante no Centre for Vision, Speech and Signal Processing (CVSSP), University of Surrey, onde realizou estágio pós-doutoral em 2016. Coordenador de projetos de pesquisa nas áreas de Visão Computacional e Reconhecimento de Padrões financiado por agências públicas (FAPESP, CNPq), institutos e empresas (Google, UGPN). Autor de mais de 40 artigos publicados em conferências e periódicos com revisão por pares. Atua nas áreas de Processamento de Sinais, Imagens e Videos, com ênfase em problemas inversos, aprendizado e extração de características com aplicações.

\begin{wrapfigure}{l}{0.2\linewidth}
 \includegraphics[width=1\linewidth]{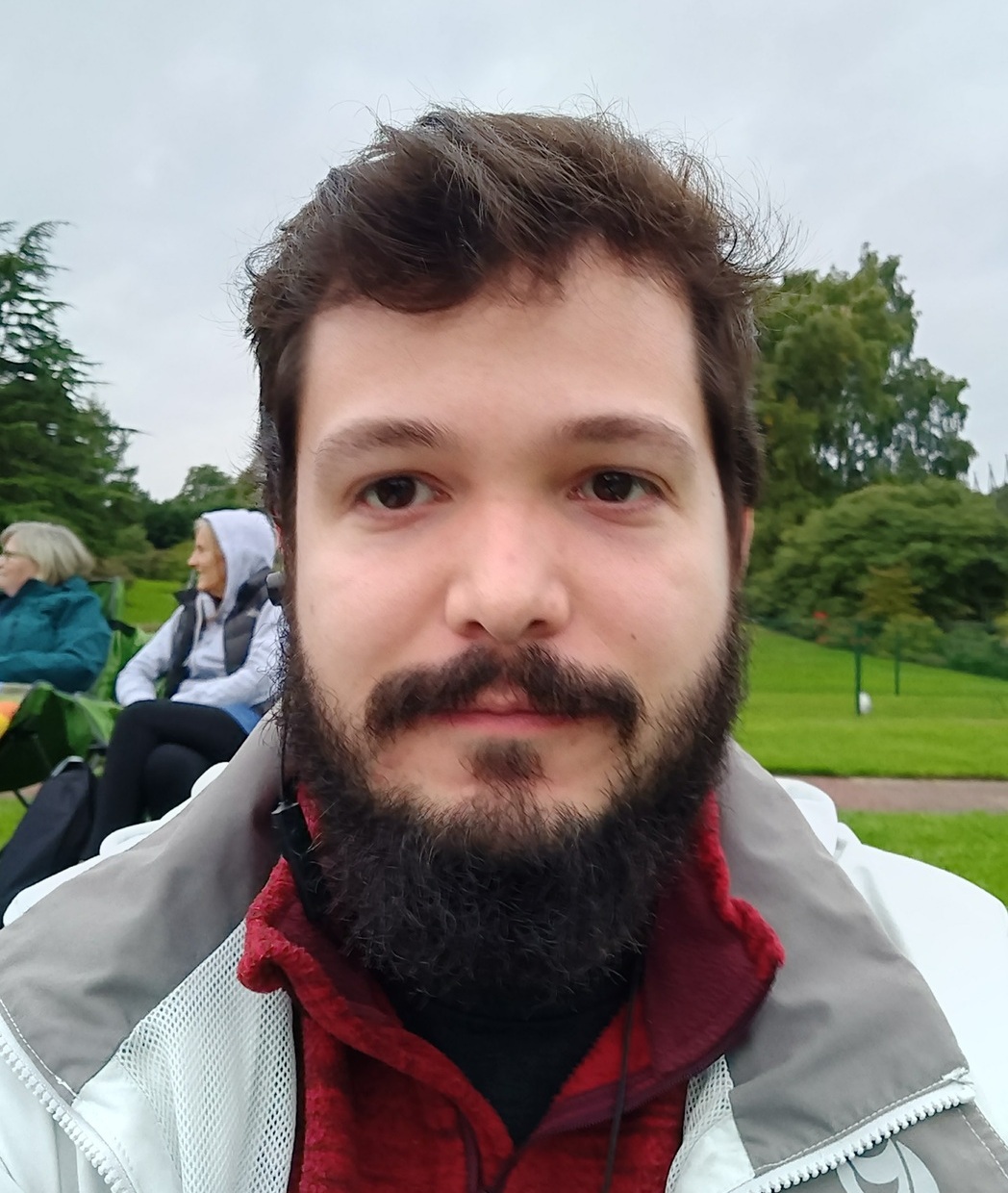}
\end{wrapfigure}
\paragraph{Gabriel B. Paranhos da Costa} é doutorando no Instituto de Ciências Matemáticas e de Computação (ICMC) da Universidade de São Paulo (USP), em São Carlos/SP, Brasil, com período sanduíche na Universidade de Edimburgo, Reino Unido. Possui Bacharelado em Ciências da Computação (2012) e Mestrado (2014) pelo ICMC/USP. Realiza pesquisas na área de Processamento de Imagens e Reconhecimento de Padrões, tendo como principal foco detecção de anomalias e aprendizado e extração de características para aplicações de aprendizado de máquina.

% that's all folks
\end{document}